\documentclass[acmsmall]{acmart}

\AtBeginDocument{%
  \providecommand\BibTeX{{%
    \normalfont B\kern-0.5em{\scshape i\kern-0.25em b}\kern-0.8em\TeX}}}

\usepackage{xspace}
\usepackage{amsfonts}
\usepackage{algorithm, algorithmic}

\DeclareMathOperator*{\argmax}{argmax}
\DeclareMathOperator*{\argmin}{argmin}
\newcommand{\name}{{ApproxNet}\xspace}
\newcommand{\eg}{{e.g.,}\xspace}
\newcommand{\ie}{{i.e.,}\xspace}
\newcommand{\et}{\textit{et al.}\xspace}
\newcommand{\iconsupport}{\includegraphics[width=0.17in]{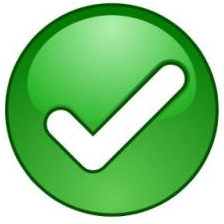}}
\newcommand{\iconpartsupport}{\includegraphics[width=0.17in]{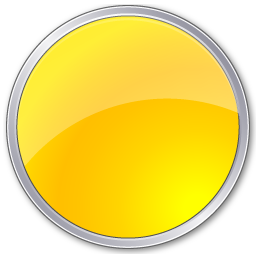}}
\newcommand{\iconnotsupport}{\includegraphics[width=0.17in]{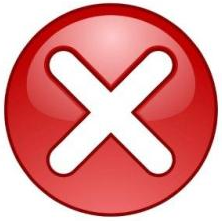}}

\begin{document}

\title{ApproxNet: Content and Contention-Aware Video Object Classification System for Embedded Clients}

\author{Ran Xu}
\affiliation{%
  \institution{Purdue University}
  \streetaddress{610 Purdue Mall}
  \city{West Lafayette}
  \state{Indiana}
  \country{USA}
  \postcode{47907}}
\email{xu943@purdue.edu}

\author{Rakesh Kumar}
\affiliation{%
  \institution{Microsoft Corp}
  \streetaddress{One Microsoft Way}
  \city{Redmond}
  \state{Washington}
  \country{USA}
  \postcode{98052}}
\email{rakku@microsoft.com}

\author{Pengcheng Wang}
\affiliation{%
  \institution{Purdue University}
  \country{USA}}
\email{wang4495@purdue.edu}

\author{Peter Bai}
\affiliation{%
  \institution{Purdue University}
  \country{USA}}
\email{pbai@purdue.edu}

\author{Ganga Meghanath}
\affiliation{%
  \institution{Indian Institute of Technology Madras}
  \streetaddress{Indian Institute Of Technology}
  \city{Chennai}
  \state{Tamil Nadu}
  \country{India}
  \postcode{600036}}
\email{gangamegha29@gmail.com}

\author{Somali Chaterji}
\affiliation{%
  \institution{Purdue University}
  \country{USA}}
\email{schaterji@purdue.edu}

\author{Subrata Mitra}
\affiliation{%
  \institution{Adobe Research}
  \streetaddress{345 Park Avenue}
  \city{San Jose}
  \state{California}
  \country{USA}
  \postcode{95110-2704}
}
\email{subrata.mitra@adobe.com}

\author{Saurabh Bagchi}
\affiliation{%
  \institution{Purdue University}
  \country{USA}}
\email{sbagchi@purdue.edu}

\renewcommand{\shortauthors}{Xu \et}

\begin{abstract}

Videos take a lot of time to transport over the network, hence running analytics on the live video on embedded or mobile devices has become an important system driver. Considering that such devices, \eg surveillance cameras or AR/VR gadgets, are resource constrained, creating lightweight deep neural networks (DNNs) for embedded devices is crucial. None of the current approximation techniques for object classification DNNs can adapt to changing runtime conditions, \eg changes in resource availability on the device, the content characteristics, or requirements from the user. In this paper, we introduce \name, a video object classification system for embedded or mobile clients. It enables novel dynamic approximation techniques to achieve desired inference latency and accuracy trade-off under changing runtime conditions. It achieves this by enabling two approximation knobs within a single DNN model, rather than creating and maintaining an ensemble of models (\eg MCDNN [MobiSys-16]. We show that \name can adapt seamlessly at runtime to these changes, provides low and stable latency for the image and video frame classification problems, and show the improvement in accuracy and latency over ResNet [CVPR-16], MCDNN [MobiSys-16], MobileNets [Google-17], NestDNN [MobiCom-18], and MSDNet [ICLR-18].

\end{abstract}



\maketitle

\section{Introduction}
\label{sec_introduction}

There is an increasing number of scenarios where various kinds of analytics are required to be run on live video streams, on resource-constrained mobile and embedded devices. For example, in a smart city traffic system, vehicles are redirected by detecting congestion from the live video feeds from traffic cameras while in Augmented Reality (AR)/Virtual Reality (VR) systems, scenes are rendered based on the recognition of objects, faces or actions in the video. These applications require low latency for event classification or identification based on the content in the video frames. Most of these videos are captured at end-client devices such as IoT devices, surveillance cameras, or head-mounted AR/VR systems. Video transport over wireless network is slow and these applications often must operate under intermittent network connectivity. Hence such systems must be able to run video analytics in-place, on these resource-constrained client devices\footnote{For end client devices, we will use the term ``mobile devices'' and ``embedded devices'' interchangeably. The common characteristic is that they are computationally constrained. While the exact specifications vary across these classes of devices, both are constrained enough that they cannot run streaming video analytics without approximation techniques.} to meet the low latency requirements for the applications. 

\noindent\textbf{State-of-the-art is too heavy for embedded devices:}
Most of the video analytics queries involve performing an inference over DNNs (mostly convolutional neural networks, {\em aka} CNNs) with a variety of functional architectures for performing the intended tasks like classification~\cite{simonyan2014very, szegedy2015going, resnet, huang2017densely}, object detection~\cite{liu2016ssd, redmon2016you, ren2015faster, shankar2020janus}, face~\cite{parkhi2015deep, schroff2015facenet, wen2016discriminative, taigman2014deepface} or action recognition~\cite{ji20133d, poppe2010survey, liu2016spatio, simonyan2014two} etc. With advancements in deep learning and emergence of complex architectures, DNN-based models have become \textit{deeper} and {\em wider}. Correspondingly their memory footprints and their inference latency have become significant. For example, DeepMon~\cite{deepmon} runs the VGG-16 model at approximately 1-2 frames-per-second (fps) on a Samsung Galaxy S7. ResNet~\cite{resnet}, with its 101-layer version, has a memory footprint of 2.8 GB and takes 101 ms to perform inference on a single video frame on the NVIDIA Jetson TX2. MCDNN, Mainstream, VideoStorm and Liu \et~\cite{mcdnn, jiang2018mainstream, liu2019edge, videostorm} require either the cloud or the edge servers to achieve satisfactory performance. Thus,  on-device inference with a low and stable inference latency (\ie 30 fps) remains a challenging task.

\begin{figure}[b]
  \centering
  \begin{minipage}{0.49\columnwidth}
    \begin{minipage}{0.49\columnwidth}
      \centering
      \includegraphics[width=1\textwidth]{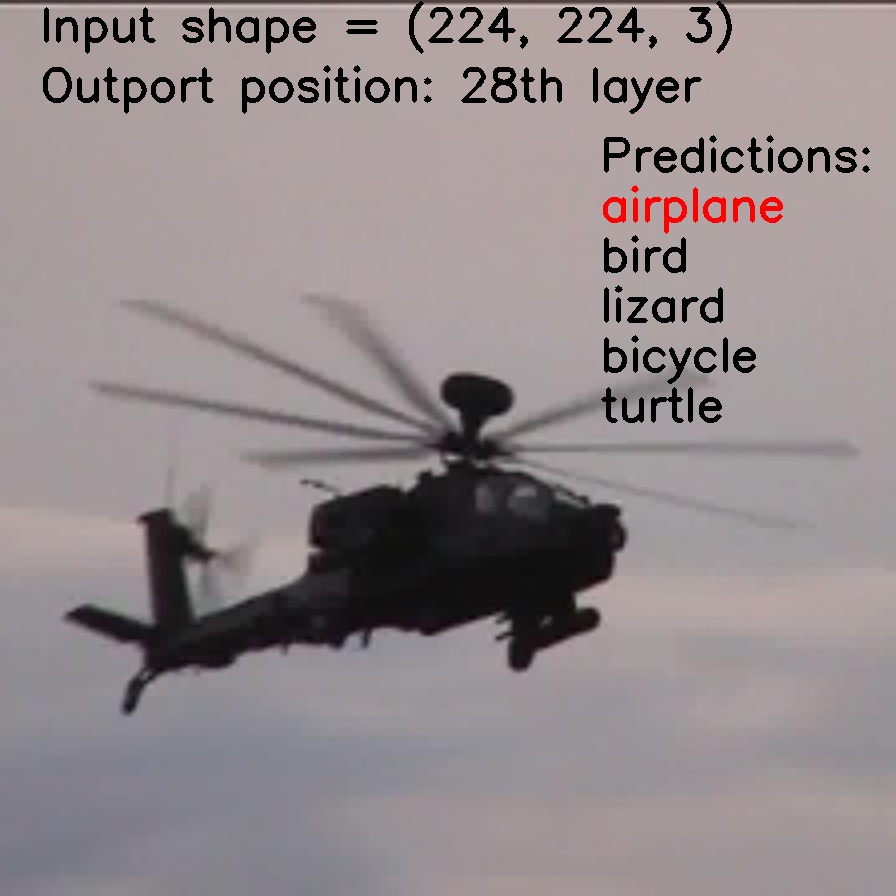}
    \end{minipage}
    \centering\hfill
    \begin{minipage}{0.49\columnwidth}
      \centering
      \includegraphics[width=1\textwidth]{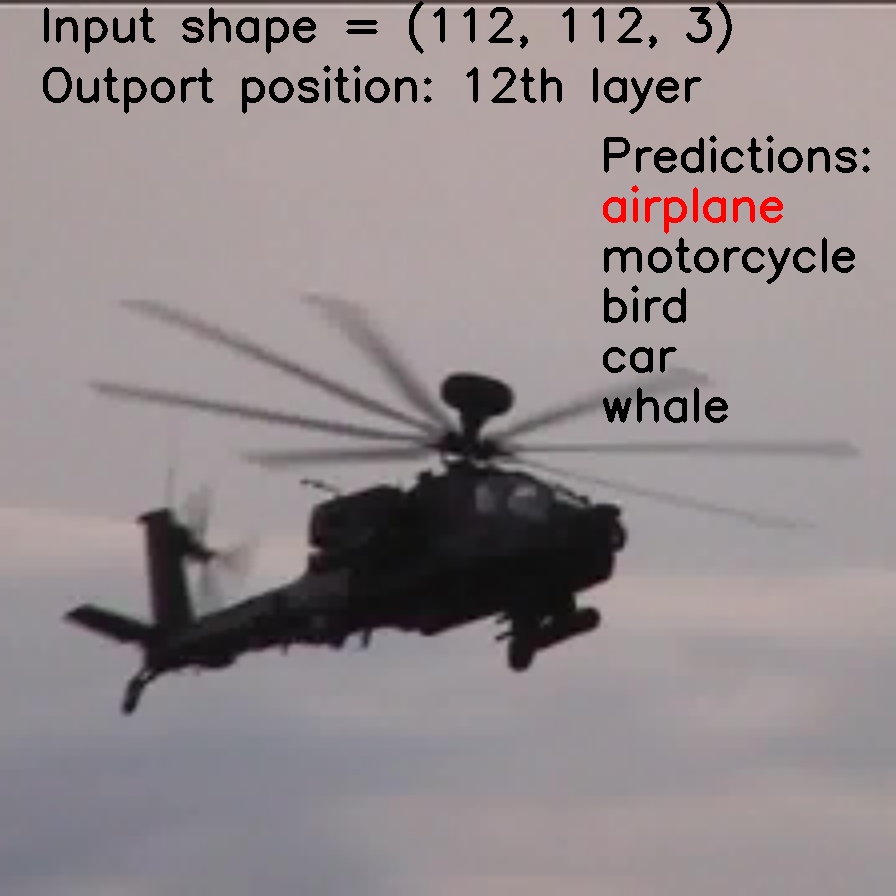}
    \end{minipage}
    \centerline{(a) Simple video frame}
  \end{minipage}
  \begin{minipage}{0.49\columnwidth}
    \begin{minipage}{0.49\columnwidth}
      \centering
      \includegraphics[width=1\textwidth]{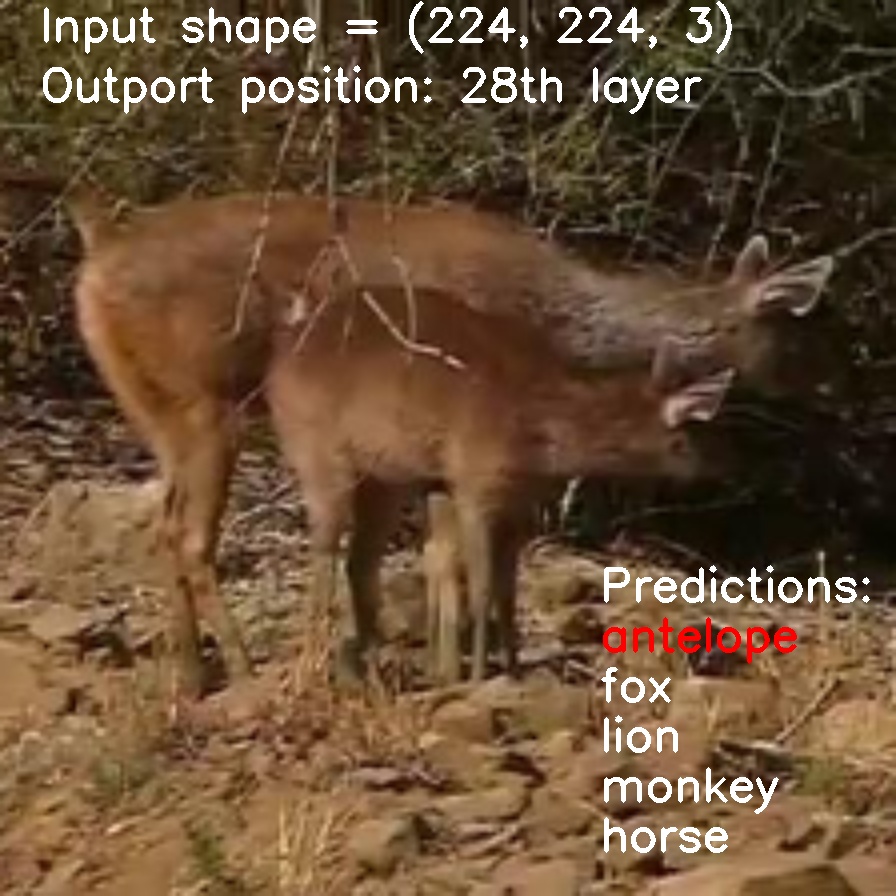}
    \end{minipage}
    \centering\hfill
    \begin{minipage}{0.49\columnwidth}
      \centering
      \includegraphics[width=1\textwidth]{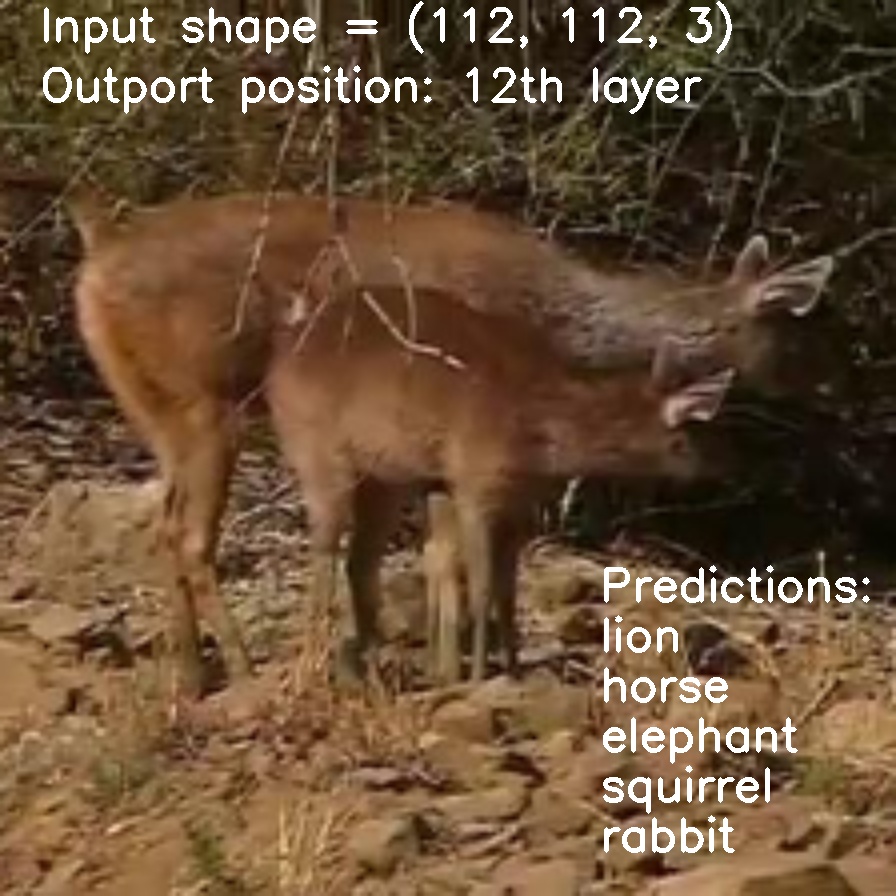}
    \end{minipage}
  \centerline{(b) Complex video frame}
  \end{minipage}
  \caption{Examples of using a heavy DNN (on the left) and a light DNN (on the right) for simple and complex video frames in a video frame classification task. The light DNN downsamples an input video frame to half the default input shape and gets prediction labels at an earlier layer. The classification is correct for the simple video frame (red label denotes the correct answer) but not for the complex video frame.}
  \label{fig:easy_hard}
\end{figure}

\noindent\textbf{Content and contention aware systems: } Content characteristics of the video stream is one of the key runtime conditions of the systems with respect to approximate video analytics. This can be leveraged to achieve the desired latency-accuracy tradeoff in the systems. For example, as shown in Figure~\ref{fig:easy_hard}, if the frame is very simple, we can downsample it to half of the original dimensions and use a shallow or small DNN model to make an accurate prediction. If the frame is complex, the same shallow model might result in a wrong prediction and would need a larger DNN model.

Resource contention is another important runtime condition that the video analytics system should be aware of. In several scenarios, the mobile devices support multiple different applications, executing concurrently. For example, while an AR application is running, a voice assistant might kick in if it detects background conversation, or a spam filter might become active, if emails are received. All these applications share common resources on the device, such as, CPU, GPU, memory, and memory bandwidth, and thus lead to \textit{resource contention}~\cite{kayiranMICRO2014, ausavarungnirunASPLOS2018, bagchi2020new} as these devices do not have advanced resource isolation mechanisms. It is currently an unsolved problem how video analytics systems running on the mobile devices can maintain a low inference latency under such variable resource availability and changing content characteristics, so as to deliver satisfactory user experience.

\noindent\textbf{Single-model vs. multi-model adaptive systems}: How do we architect the system to operate under such varied runtime conditions?  Multi-model designs came first in the evolution of systems in this space. These created systems with an ensemble of multiple models, satisfying varying latency-accuracy conditions, and some scheduling policy to choose among the models. MCDNN~\cite{mcdnn}, being one of most representative works, and well-known DNNs like ResNet, MobileNets~\cite{resnet, howard2017mobilenets} all fall into this category.  On the other hand, the single-model designs, which emerged after the multi-model designs, feature one model with tuning knobs inside so as to achieve different latency-accuracy goals. These typically have lower switching overheads from one execution path to another, compared to the multi-branch models.  MSDNet~\cite{huang2017multi}, BranchyNet~\cite{teerapittayanon2016branchynet}, NestDNN~\cite{fang2018nestdnn} are representative works in this single-model category. However, none of these systems can adapt to runtime conditions, primarily, changes in content characteristics and contention levels on the device.



\begin{table}[b]
  \centering
  \caption{ApproxNet's main features and comparison to existing systems.}
  \label{table:features}
  \scalebox{0.85}{
  \begin{tabular}{|p{1.4in}|p{0.4in}|p{0.8in}|p{0.5in}|p{0.6in}|p{0.5in}|p{0.6in}|}
    \hline
    Solution & Single model & Considers switching overhead & Focused on video & Handles runtime conditions & Open-sourced & Replicable in our datasets\\
    \hline
    MCDNN [MobiSys'16]      & \iconnotsupport & \iconpartsupport & \iconnotsupport& \iconnotsupport & \iconpartsupport & \iconsupport  \\
    \hline
    MobileNets [ArXiv'17] & \iconnotsupport & \iconnotsupport & \iconnotsupport& \iconnotsupport & \iconsupport & \iconsupport \\
    \hline
    MSDNet [ICLR'18]     & \iconsupport & \iconnotsupport & \iconnotsupport& \iconnotsupport & \iconsupport & \iconsupport \\
    \hline
    BranchyNet [ICPR'16] & \iconsupport & \iconnotsupport & \iconnotsupport& \iconnotsupport & \iconsupport & \iconnotsupport  \\
    \hline
    NestDNN [MobiCom'18] & \iconsupport & \iconpartsupport & \iconnotsupport& \iconpartsupport & \iconnotsupport & \iconnotsupport \\
    \hline
    ApproxNet            & \iconsupport & \iconsupport & \iconsupport      & \iconsupport  & \iconsupport & \iconsupport \\
    \hline
    \multicolumn{7}{|c|}{\iconsupport Supported \iconpartsupport Partially Supported \iconnotsupport Not Supported} \\
    \hline
    \multicolumn{7}{|l|}{Notes for partially support:} \\
    \multicolumn{7}{|l|}{1. MCDNN and NestDNN only consider the switching overhead in memory size, but in the execution latency.} \\
    \multicolumn{7}{|l|}{2. NestDNN handles multiple concurrent DNN applications with joint optimization goals.} \\
    \multicolumn{7}{|l|}{3. The core models in MCDNN are open-sourced while the scheduling components are not.} \\
    \hline
  \end{tabular}
  }
\end{table}

\noindent {\bf Our solution: \name}. 
In this paper, we present \name, our content and contention aware object classification system over streaming videos, geared toward GPU-enabled mobile/embedded devices. We introduce a novel workflow with a set of integrated techniques to solve the three main challenges as mentioned above: (1) on-device real-time video analytics, (2) content and contention aware runtime calibration, (3) a single-model design. The fundamental idea behind \name is to perform approximate computing with tuning knobs that are changed automatically and seamlessly within the same video stream. These knobs trade off the accuracy of the inferencing for reducing inference latency and thus match the frame rate of the video stream or the user's requirement on either a latency target or an accuracy target. The optimal configuration of the knobs is set, particularly considering the resource contention and complexity of the video frames, because these runtime conditions affects the accuracy and latency of the model much. 

In Table~\ref{table:features}, we compare \name with various representative prior works in this field. First of all, none of these systems~\cite{mcdnn, howard2017mobilenets, huang2017multi, teerapittayanon2016branchynet, fang2018nestdnn} is able to adapt to dynamic runtime conditions (changes in content characteristics and contention levels) as we can. Second, although most systems are able to run at variable operation points of performance, MCDNN~\cite{mcdnn} and MobileNets~\cite{howard2017mobilenets} use a multi-model approach and incur high switching penalty. For those that works with a single model, namely, MSDNet~\cite{huang2017multi}, BranchyNet~\cite{teerapittayanon2016branchynet}, and NestDNN~\cite{fang2018nestdnn}, they do not consider switching overheads in their models (except partially for NestDNN, which considers switching cost in memory size), do not focus on video content, and do not show how their models can adapt to the changing runtime conditions (except partially for NestDNN, which considers joint optimization of multiple DNN workloads). For evaluation, we mainly compare to MCDNN, ResNet, and MobileNets, as the representatives of multi-model approaches, and MSDNet, as the single-model approach. We cannot compare to BranchyNet as it is not designed and evaluated for video analytics and thus not suitable for our datasets. BranchyNet paper evaluates it on small image dataset: MNIST and CIFAR. We cannot compare to NestDNN since it's models or source-code and architecture and hyperparameter details are publicly available and we need those for replicating the experiments. 

To summarize, we make the following contributions in this paper:
\begin{enumerate}

\item We develop an end-to-end, approximate video object classification system, \name, that can handle dynamically changing workload contention and video content characteristics on resource-constrained embedded devices. It achieves this through performing system context-aware and content-aware approximations with the offline profiling and online lightweight sensing and scheduling techniques.

\item We design a novel workflow with a set of integrated techniques including the adaptive DNN that allows runtime accuracy and latency tuning \textit{within a single model}. Our design is in contrast to ensemble systems like MCDNN that are composed of multiple independent model variants capable of satisfying different requirements. Our single-model design avoids high switching latency when conditions change and reduces RAM and storage usage.

\item We design \name to make use of video features, rather than treating video as a sequence of image frames. Such characteristics that we leverage include the temporal continuity in content characteristics between adjacent frames. We empirically show that on a large-scale video object classification dataset, popular in the vision community, \name achieves a superior accuracy-latency tradeoff than the three state-of-the-art solutions on mobile devices, MobileNets, MCDNN, and MSDNet (Figures~\ref{trade_off_img_dataset} and \ref{trade_off_system}).

\end{enumerate}

The rest of the paper is organized as follows. Section~\ref{sec_background} gives the relevant background. Section~\ref{sec_overview} gives our high-level solution overview. Section~\ref{sec_technique} gives the detailed design. Section~\ref{sec_evaluation} evaluates our end-to-end system. Section~\ref{sec_discussion} discusses the details about training the DNN. Section~\ref{sec_related_work} highlights the related works. Finally, Section~\ref{sec_conclusion} gives concluding remarks.

\section{Background and Motivation}
\label{sec_background}

\subsection{DNNs for Streaming Video Analytics}

DNNs have become a core element of various video processing tasks such as frame classification, human action recognition, object detection, face recognition, and so on. Though accurate, DNNs are computationally expensive, requiring significant CPU and memory resources.  As a result, these DNNs are often too slow when running on mobile devices and become the latency bottleneck in video analytics systems. Huynh \et~\cite{deepmon} experimented with VGG~\cite{simonyan2014very} of 16 layers on the Samsung Galaxy S7 and noted that classification on a single image takes as long as 644 ms, leading to less than 2 fps for continuous classification. Motivated by the observation, we explore in \name how we can make DNN-based video analytics pipelines more efficient through content-aware approximate computation {\em within the neural network}.

{\noindent\bf ResNet}: 
Deep DNNs are typically hard to train due to the vanishing gradient problem~\cite{resnet}. ResNet solved this problem by introducing a short cut identity connection in between layers, which helps achieve at least the same accuracy upon further increasing the number of network layers. The unit of such connected layers is called a ResNet block. We leverage this key idea of a deeper model producing no higher error than its shallower counterpart, for the construction of an architecture that provides more accurate execution as it proceeds deeper into the DNN.

{\noindent{\bf Spatial Pyramid Pooling (SPP)}~\cite{spp}}: 
Popular DNN models, including ResNet, consist of convolutional and max-pooling (CONV) layers and fully-connected (FC) layers and the shape of an input image is fixed. Changing the input shape in a CNN typically requires re-designing the architecture of the CNN. SPP is a special layer that eliminates this constraint. The SPP layer is added at the end of CONV layers, providing the following FC layers with a fixed-dimensioned feature representation by pooling the CONV layer output with bins whose shapes are proportional to the input shape. We use SPP layers to change input shape as an approximation knob.

{\noindent\bf Trading-off accuracy for inference latency}: 
DNNs can have several variants due to different configurations, and these variants yield different accuracies and latencies. But these variants are trained and inferenced independently and cannot be switched efficiently at inference time to meet differing accuracy or latency requirements. For example, MCDNN~\cite{mcdnn} sets up an ensemble of (up to 68) model variants to satisfy different latency/accuracy/cost requirements. MSDNet~\cite{huang2017multi} enables five early exits in a {\em single} model but does not evaluate on streaming video with any variable content or contention situations. Hence, we set ourselves to design a single-model DNN that is capable of handling the accuracy-latency trade-off at inference time and guarantees our video analytics system's performance under variable content and runtime conditions.

\subsection{Content-aware Approximate Computing}

IRA~\cite{laurenzano2016input} and VideoChef~\cite{xu2018videochef} first introduced the notion of content-aware approximation and applied the idea, respectively to image and video processing pipelines. These works for the first time showed how to tune approximation knobs as content characteristics change, \eg the video scene became more complex. In particular, IRA performs approximation targeting individual images, while VideoChef exploits temporal similarity among frames in a video to further optimize accuracy-latency trade-off. However, these works do not perform approximation for ML-based inferencing, which comprises the dominant form of video analytics. In contrast, we apply approximation to the DNN model itself with the intuition that depending on complexity of the video frame, we want to feed input of a different shape and output at a different depth of layers to achieve the target accuracy.

\subsection{Contention-aware Scheduling}

Managing the resource contention of multiple jobs on high-performance clusters is a very active area of work. Bubble-Up~\cite{mars2011bubble}, Bubble-Flux~\cite{yang2013bubble}, and Pythia~\cite{xu2018pythia} develop characterization methodologies to predict the performance degradation of latency-sensitive applications due to shared resources in the memory subsystem. SMiTe~\cite{zhang2014smite} and Paragon~\cite{delimitrou2013paragon} further extend such concurrent resource contention scenario to SMT processors and thousands of different unknown applications, respectively. On the other hand, we apply contention-aware approximation to the DNN model on the embedded and mobile devices, and consider the three major sources of contention -- CPU, GPU, and memory bandwidth.

\section{Overview}
\label{sec_overview}

Here we give a high-level overview of \name. In Section~\ref{sec_technique}, we provide details of each component. 

\subsection{Design Principles and Motivation}

We set four design requirements for streaming video analytics on the embedded devices motivated by real-world scenarios and needs. {\em First}, the application should adapt to changing input characteristics, such as, the complexity of the video frames because the accuracy of the DNN may vary based on the content characteristics. We find such changes happen often enough within a single video stream and without any clear predictive pattern. {\em Second}, the application should adapt to the resource contention due to the shared CPU, GPU, memory, or memory bandwidth with other concurrent applications on the same device. Such contention can happen frequently with co-location due to limited resources and the lack of clean resource isolation on these hardware platforms. Again, we find that such changes can happen without a clear predictive pattern. {\em Third}, the application should support different target accuracies or latencies at runtime with little transition overhead. For example, the application may require low latency when a time-critical query, such as detection of a miscreant, needs to be executed and have no such constraint for other queries on the steam. Thus, the aggregate model must be able to make efficient transitions in the tradeoff space of accuracy and latency, and less obviously throughput, optionally using edge or cloud servers. {\em Fourth}, the application must provide real-time processing speed (30 fps) while running on the mobile/embedded device. To see three instances where these four requirements come together, consider mobile VR/AR games like Pokemon Go (some game consoles support multitasking, accuracy requirements may change with the context of the game), autonomous vehicles (feeds from multiple cameras are processed on the same hardware platform resulting in contention, emergency situations require lower latency than benign conditions such as for fuel efficiency) and autonomous drones (same arguments as for autonomous vehicles). 

A {\em non-requirement} in our work is that multiple concurrent applications consuming the same video stream be jointly optimized. MCDNN~\cite{mcdnn}, NestDNN~\cite{fang2018nestdnn}, and Mainstream~\cite{jiang2018mainstream} bring significant design sophistication to handle the concurrency aspect. However, we are only interested in optimizing a single video analytics application. 

\subsection{Design Intuition and Workflow}
\label{subsec:workflow}

\begin{figure}[t]
  \centering
   \includegraphics[width=0.99\textwidth]{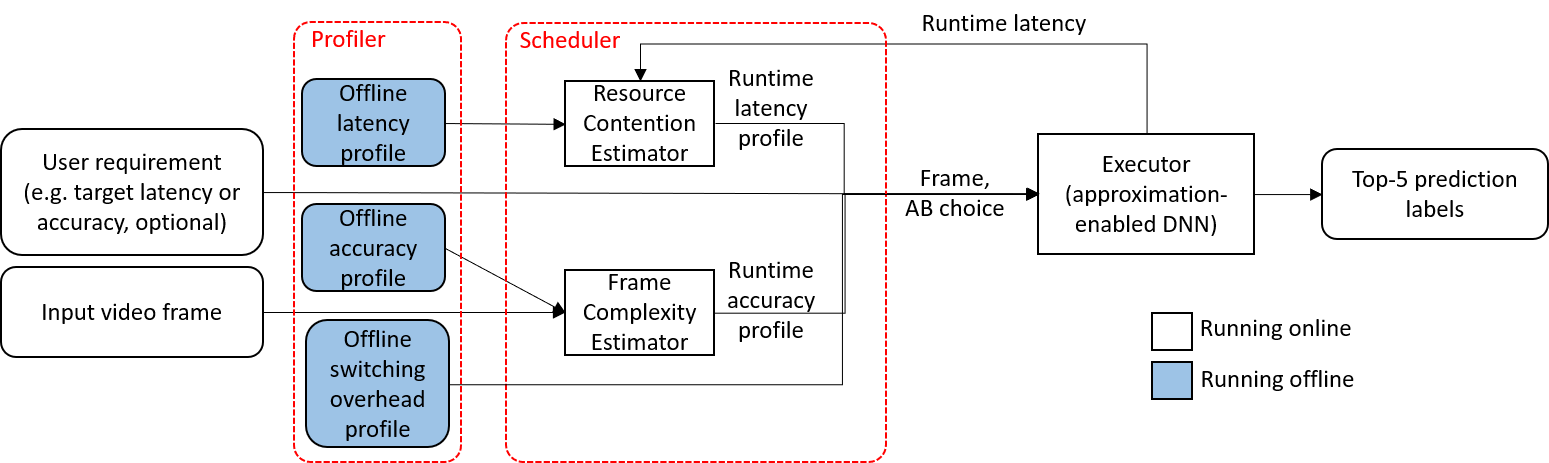}
   \caption{Workflow of \name. The input is a video frame and an optional user requirement, and the outputs are prediction labels of the video frame. Note that the shaded profiles are collected offline to alleviate the online scheduler overhead.}
   \label{fig:overview}
\end{figure}

To address these challenges in our design, we propose a novel workflow with a set of integrated techniques to achieve a content and contention-aware video object classification system. We show the overall structure with three major functional units: \textit{executor}, \textit{profiler}, and \textit{scheduler} in Figure~\ref{fig:overview}. \name takes a video frame and optional user requirement for target accuracy or latency as an input, and produces top-5 prediction labels of the object classes as outputs.

The executor (Section~\ref{sec:approx_dnn}) is an approximation-enabled, single-model DNN. Specifically, the single-model design largely reduces the switching overhead so as to support the adaptive system. On the other hand, the multiple \textbf{approximation branches (ABs)}, each with variable latency and accuracy specs, are the key to support dynamic content and contention conditions. \name is designed to provide real-time processing speed (30 fps) on our target device (NVIDIA Jetson TX2). Compared to the previous single-model designs like MSDNet and BranchyNet, \name provides novelty in enabling both depth and shape as the approximation knob for run-time calibration. 

The scheduler is the key component to react to the dynamic content characteristics and resource contention. Specifically, it selects an AB to execute by combining the precise accuracy estimation of each AB due to changing content characteristics via a {\bf Frame Complexity Estimator} ({\bf FCE}, Section~\ref{subsec_FCE}), the precise latency estimation of each AB due to resource contention via a {\bf Resource Contention Estimator} ({\bf RCE}, Section~\ref{subsec_RCE}), and the switching overhead among ABs (Section~\ref{sec:profiler}). It finally reaches a decision on which AB to use based on the user's latency or accuracy requirement and its internal accuracy, latency, and overhead estimation (Section~\ref{subsec:pareto}).

Finally, to achieve our goal of real-time processing, low switching overhead, and improved performance under dynamic conditions, we design an offline profiler (Section~\ref{sec:profiler}). We collect three profiles offline --- first, the accuracy profile for each AB on video frames of different complexity categories; second, the inference latency profile for each AB under variable resource contention, and third, the switching overhead between any two ABs.

{\bf Video-specific design}. We incorporate these video-specific designs in \name, which is orthogonal to the existing techniques presented in the prior works on video analytics, e.g. frame sampling~\cite{jiang2018mainstream, videostorm} and edge device offloading~\cite{liu2019edge}.

\begin{enumerate}

\item
The FCE uses a Scene Change Detector (SCD) as a preprocessing module to further alleviate its online cost. This optimization is beneficial because it reduces the frequency with which the FCE is invoked (only when the SCD flags a scene change). This relies on the intuition that discontinuous jumps in frame complexity are uncommon in a video stream.

\item The scheduler decides whether to switch to a new AB or stay depending on how long it predicts the change in the video stream to last and the cost of switching. 

\item We drive our decisions about the approximation knobs by the goal of keeping up with the streaming video rate (30 fps). We achieve this under most scenarios when evaluated with a comprehensive video dataset (the ILSVRC VID dataset).

\end{enumerate}
\section{Design and Implementation}
\label{sec_technique}

\subsection{Approximation-enabled DNN}
\label{sec:approx_dnn}

\name's key enabler, an approximation-enabled DNN, is designed to support multiple accuracy and latency requirements at runtime \textit{using a single DNN model}. To enable this, we design a DNN that can be approximated using two approximation knobs. The DNN can take an input video frame in different shapes, which we call \textit{input shapes}, our first approximation knob and it can produce a classification output at multiple positions in the intervening layers, which we call \textit{outports}, our second approximation knob. There are doubtless other approximation knobs, \eg model quantization, frame sampling, and others depending on specific DNN models. These can be incorporated into \name and they will all fit within our novelty of the one-model DNN to achieve real-time on-device, adaptive inference. The appropriate setting of the tuning knobs can be determined on the device (as is done in our considered usage scenario) or, in case this computation is too heavyweight, this  can be determined remotely and sent to the device through a reprogramming mechanism such as~\cite{panta2011efficient}.

\begin{figure*}[t]
  \begin{minipage}[t!]{0.35\linewidth} 
    \centering
    \includegraphics[width=1\columnwidth]{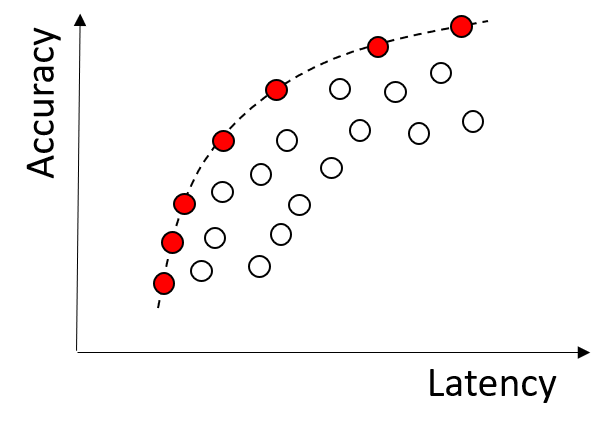}
    \caption{A Pareto frontier for trading-off accuracy and latency in a particular frame complexity category and at a particular contention level.}
    \label{fig:pareto}
  \end{minipage} 
  \hfill
  \begin{minipage}[t!]{0.60\linewidth} 
    \centering
    \includegraphics[width=1\columnwidth]{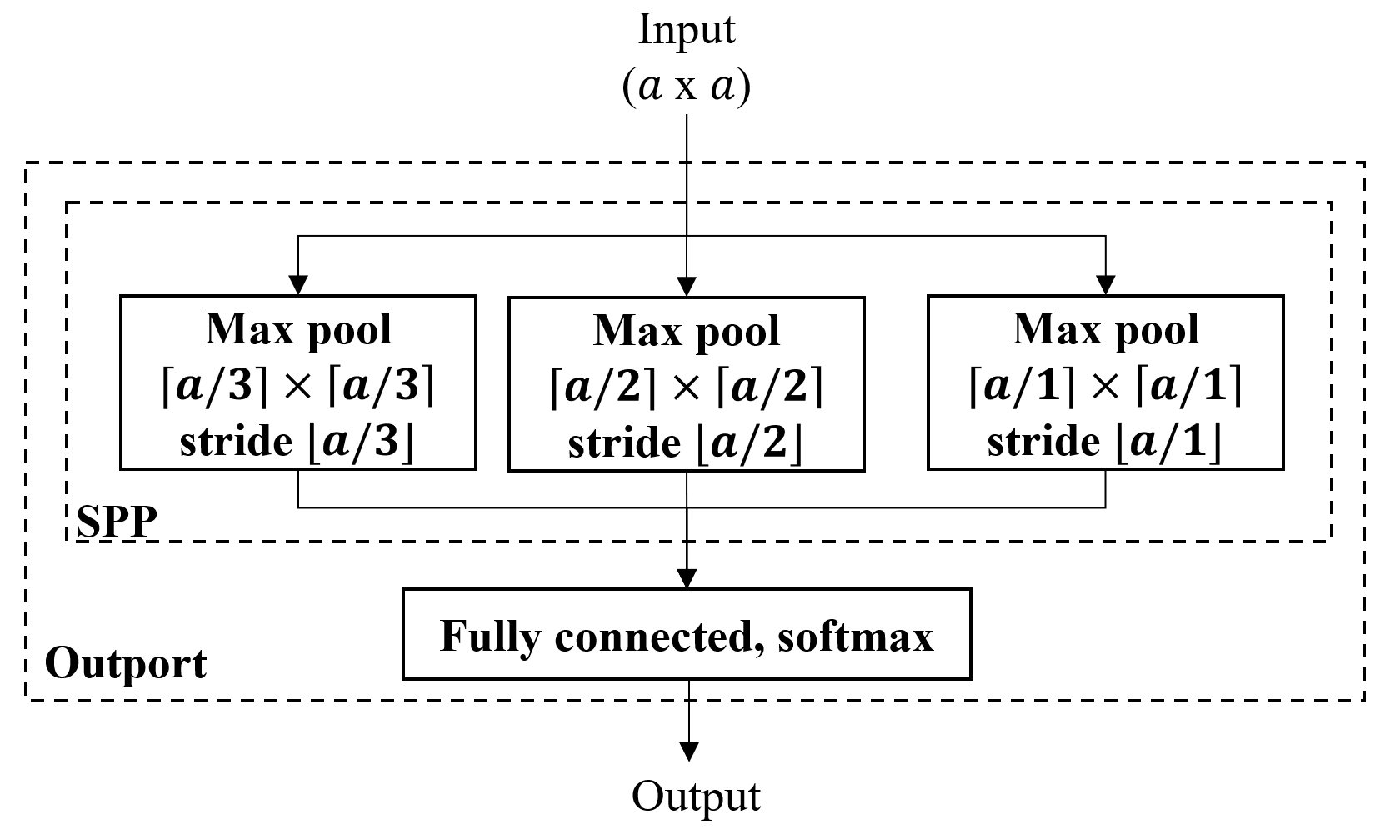}
    \caption{The outport of the approximation-enabled DNN.}
    \label{fig:detailed_DNN}
  \end{minipage}
\end{figure*}

Combining these two approximation knobs, \name creates various {\bf approximation branches (ABs)}, which trade off between accuracy and latency, and can be used to meet a particular user requirement. This tradeoff space defines a set of Pareto optimal frontiers, as shown in Figure~\ref{fig:pareto}. Here, the scatter points represent the accuracy and latency achieved by all ABs. A Pareto frontier defines the ABs which are either superior in accuracy or latency against all other branches. 

\begin{figure*}[t]
    \centering
    \includegraphics[width=0.8\columnwidth]{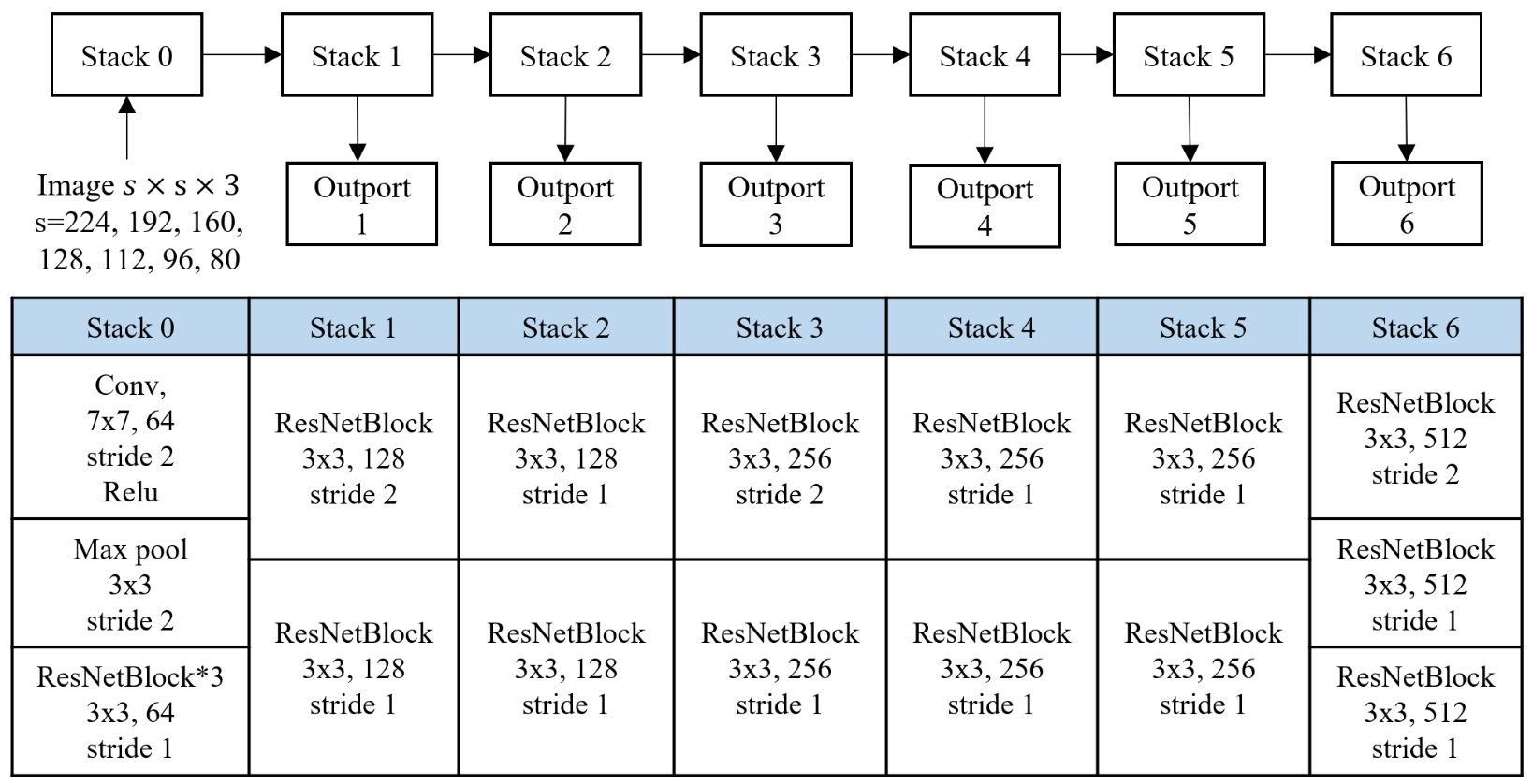}
    \caption{The architecture of the approximation-enabled DNN in \name.}
    \label{fig:dnn_arch}
\end{figure*} 

\begin{table}[t]
  \centering
  \caption{The list of the total 30 ABs supported for a baseline DNN of ResNet-34, given by the combination of the input shape and the outport from which the result is taken. ``--'' denotes the undefined settings.}
  \label{table:a_value}
  \scalebox{0.8}{
      \begin{tabular}{p{0.75in}|cccccc}
        Input shape & Outport 1 & Outport 2 & Outport 3 & Outport 4 & Outport 5 & Outport 6 \\
        \hline
        224x224x3   & 28x28x64  & 28x28x64  & 14x14x64  & 14x14x64  & 14x14x64  & 7x7x64    \\
        192x192x3   & 24x24x64  & 24x24x64  & 12x12x64  & 12x12x64  & 12x12x64  &   --      \\
        160x160x3   & 20x20x64  & 20x20x64  & 10x10x64  & 10x10x64  & 10x10x64  &   --      \\
        128x128x3   & 16x16x64  & 16x16x64  & 8x8x64    & 8x8x64    & 8x8x64    &   --      \\
        112x112x3   & 14x14x64  & 14x14x64  & 7x7x64    & 7x7x64    & 7x7x64    &   --      \\
        96x96x3     & 12x12x64  & 12x12x64  &   --      &   --      &   --      &   --      \\
        80x80x3     & 10x10x64  & 10x10x64  &   --      &   --      &   --      &   --      \\
      \end{tabular}
  }
\end{table}

We describe our design using ResNet as the base DNN, though our design is applicable to any other mainstream CNN consisting of convolutional (CONV) layers and fully-connected (FC) layers such as VGG~\cite{simonyan2014very}, DenseNet~\cite{huang2017densely} and so on. Figure~\ref{fig:dnn_arch} shows the design of our DNN using ResNet-34 as the base model. This enables 7 input shapes (${s\times s\times 3}$ for $s=224,192,160,$ $128,112,96,80$) and 6 outports (after 11, 15, 19, 23, 27, and 33 layers). We adapt the design of ResNet in terms of the stride, shape, number of channels, use of convolutional layer or maxpool, and connection of the layers. In addition, we create {\em stacks}, with stacks numbering 0 through 6 and each stack having 4 or 6 ResNet layers and a variable number of blocks from the original ResNet design (Table~\ref{table:a_value}). We then design an outport (Figure~\ref{fig:detailed_DNN}), and connect with stacks 1 to 6, whereby we can get prediction labels by executing only the stacks (i.e., the constituent layers) till that stack. The use of 6 outports is a pragmatic system choice---too small a number does not provide enough granularity to approximate in a content and contention-aware manner and too many leads to a high training burden. Further, to allow the approximation knob of downsampling the input frame to the DNN, we use the SPP layer at each outport to pool the feature maps of different shapes (due to different input shapes) into one unified shape and then connect with an FC layer. The SPP layer performs max-pooling on its input by three different levels $l=1,2,3$ with window size $\lceil a/l\rceil$ and stride $\lfloor a/l\rfloor$, where $a$ is the shape of the input to the SPP layer. Note that our choice of the 3-level pyramid pooling is a typical practice for using the SPP layer~\cite{spp}. In general, a higher value of $l$ requires a larger value of $a$ on the input of each outport, thereby reducing the number of possible ABs. On the other hand, a smaller value of $l$ results in coarser representations of spatial features and thus reduces accuracy. To support the case $l=3$ in the SPP, we require that the input shape of an outport be no less than 7 pixels in width and height, \textit{i.e}., $a\ge7$. This results in ruling out some input shapes as in Table~\ref{table:a_value}. Our model has 30 configuration settings in total, instead of 7 $\times$ 6 (number of input shapes $\times$ number of outports) because too small input shapes cannot be used when the outport is deep.

To train \name towards finding the optimal parameter set $\theta$, we consider the softmax loss $L_{s,i}(\theta)$ defined for the input shape $s\times s\times 3$ and the outport $i$. The total loss function $L(\theta)$ that we minimize to train \name is a weighted average of $L_{s,i}(\theta)$ for all $s$ and $i$, defined as 
\begin{equation}
    L(\theta)=\sum_{\forall i} \frac{1}{n_i} \sum_{\forall s} L_{s,i}(\theta)
\end{equation}
where the value of $n_i$ is the factor that normalizes the loss at an outport by dividing by the number of shapes that are supported at that port $i$. This makes each outport equally important in the total loss function. For mini-batch, we use 64 frames for each of the 7 different shapes. 
To train on a particular dataset or generalize to other architectures, we discuss more details in Section~\ref{sec_discussion}.

\begin{figure*}[t]
    \centering
    \includegraphics[width=0.7\columnwidth]{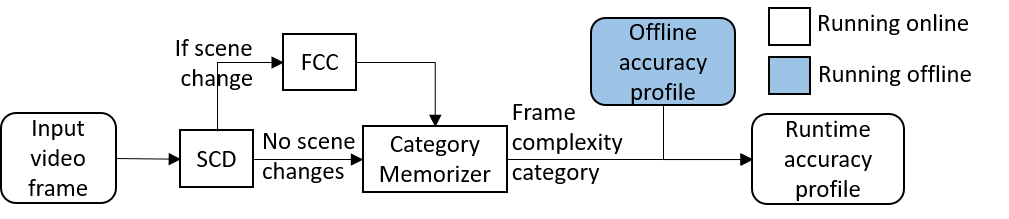}
    \caption{Workflow of the Frame Complexity Estimator.}
    \label{fig:fce}
\end{figure*}

\begin{figure*}[t]
  \begin{minipage}[thb]{0.54\linewidth}
    \includegraphics[width=1\textwidth]{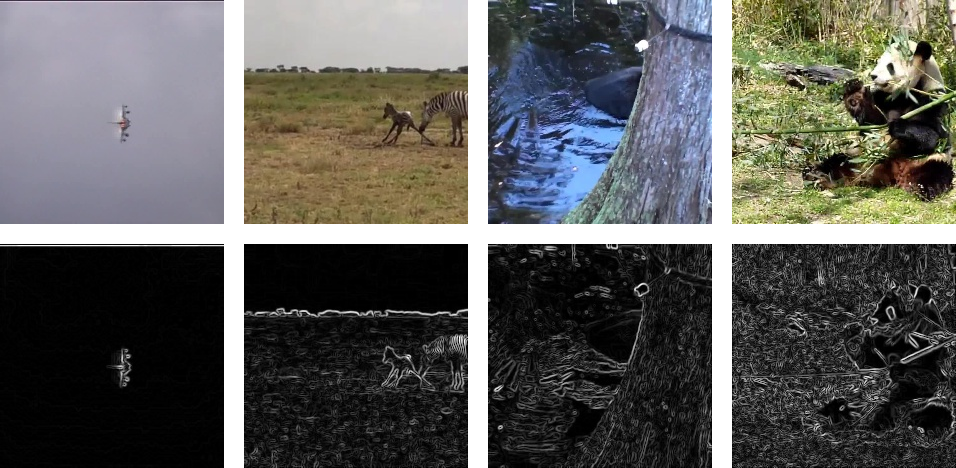}
    \caption{Sample frames (first row) and edge maps (second row), going from left to right as simple to complex. Normalized mean edge values from left to right: 0.03, 0.24, 0.50, and 0.99 with corresponding frame complexity categories: 1, 3, 6, and 7}
    \label{fig:sample_edge_maps}  
  \end{minipage} 
  \hfill 
  \begin{minipage}[thb]{0.44\linewidth}
    \includegraphics[width=1\textwidth]{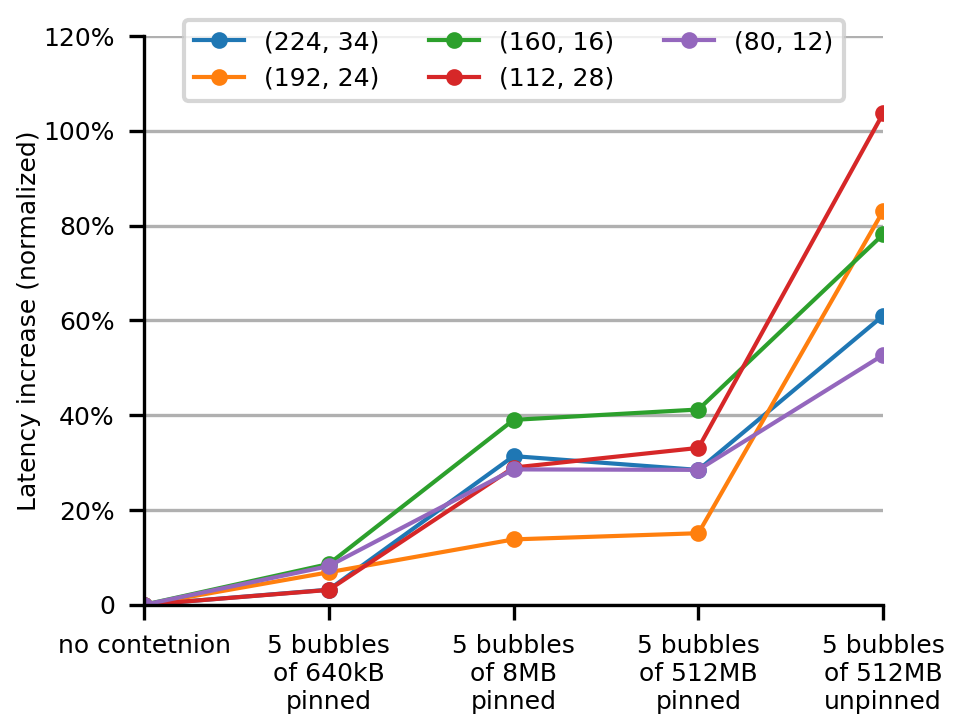}
    \caption{Latency increase of several ABs in ApproxNet under resource contention with respect to those under no contention. The input shape and outport depth of the branches are labeled.}
    \label{fig:sensitivity}  
  \end{minipage} 
\end{figure*}

\subsection{Frame Complexity Estimator (FCE)}
\label{subsec_FCE}

The design goal of the Frame Complexity Estimator (FCE), which executes online, is to estimate the expected accuracy of each AB in a content-aware manner. It is composed of a Frame Complexity Categorizer (FCC) and a Scene Change Detector (SCD) and it makes use of information collected by the offline profiler (described in Section~\ref{sec:profiler}). The workflow of the FCE is shown in Figure~\ref{fig:fce}.

\noindent \textbf{Frame Complexity Categorizer (FCC)}. FCC determines how hard it is for \name to classify a frame of the video. Various methods have been used in the literature to calculate frame complexity such as edge information-based methods~\cite{edge-compression-complexity,edge-complexity-2}, compression information-based methods~\cite{edge-compression-complexity} and entropy-based methods~\cite{cardaci2009fuzzy}. In this paper, we use mean edge value as the feature of the frame complexity category, since it can be calculated with very low computation overhead (3.9 ms per frame on average in our implementation). Although some counterexamples may show the edge value is not relevant, we show empirically that with this feature, the FCE is able to predict well the accuracy of each AB with respect to a large dataset. 

To expand, we extract an edge map by converting a color frame to a gray-scale frame, applying the Scharr operator~\cite{jahne1999handbook} in both horizontal and vertical directions, and then calculating the L2 norm in both directions. We then compute the mean edge value of the edge map and use a pre-trained set of boundaries to quantize it into several frame complexity categories. The number and boundaries of categories is discussed in Section~\ref{sec:profiler}. Figure~\ref{fig:sample_edge_maps} shows examples of frames and their edge maps from a few different complexity categories. 

\noindent \textbf{Scene Change Detector (SCD)}. The Scene Change Detector is designed to further reduce the online overhead of FCC by determining if the content in a frame is significantly different from that in a prior frame in which case the FCC will be invoked. SCD tracks a histogram of pixel values, and declares a scene change when the mean of the absolute difference across all bins of the histograms of two consecutive frames is greater than a certain threshold (45\% of the total pixels in our design). To bound the execution time of SCD we use only the R-channel and downsample the shape of the frame to $112 \times 112$. We empirically find that such optimizations do not reduce the accuracy of detecting new scenes but do reduce the SCD overhead, to only 1.3 ms per frame.

\subsection{Resource Contention Estimator (RCE)}
\label{subsec_RCE}

The design goal of the Resource Contention Estimator (RCE), which also executes online, is to estimate the expected latency of each AB in a contention-aware manner. Under resource contention, each AB is affected differently and we call the latency increase pattern the \textit{latency sensitivity}. As shown in Figure~\ref{fig:sensitivity}, five approximation branches have different ratios of latency increase under a certain amount of CPU and memory bandwidth contention. 

Ideally we would use a sample classification task to probe the system and observe its latency under the current contention level $C$. The use of such micro-benchmarks is commonly done in datacenter environments~\cite{lo2015heracles, xu2018pythia}. However, we do not need the additional probing since the inference latencies of the latest video frames form a natural observation of the contention level of the system. Thus we use the averaged inference latency $\overline{L_B}$ of the current AB $B$ across the latest $N$ frames. We then check the latency sensitivity $L_{B, C}$ of branch $B$ (offline profile as discussed in Section~\ref{sec:profiler}) and get an estimated contention level $\hat{C}$ with the nearest neighbor principle,

\begin{equation}
    \hat{C} = \argmin_C abs(L_{B, C} - \overline{L_B})
\end{equation}

By default, we use $N=30$. This will lead to an average over last one second when frame-per-second is 30. Smaller $N$ can make \name adapt faster to the resource contention, while larger $N$ make it more robust to the noise. Due to the limited observation data (one data point per frame), we cannot adapt to resource contention that is changing faster than the frame rate.

Specifically in this work, we consider CPU, GPU and memory contention among tasks executing on the device (our SoC board shares the memory between the CPU and the GPU), but our design is agnostic to what causes the contention. 
Our methodology considers the resource contention as a black-box model because we position ourselves as an application-level design instead of knowing the execution details of all other applications. We want to deal with the effect of contention, rather than mitigating it by modifying the source of the contention.

\subsection{Offline Profiler}
\label{sec:profiler}

\noindent \textbf{Per-AB Content-Aware Accuracy Profile}. 
The boundaries of frame complexity categories are determined based on the criteria that all frames within a category should have an identical Pareto frontier curve (Figure~\ref{fig:pareto}) and frames in different categories should have distinct curves. We start with considering the whole set of frames as belonging to a single category and split the range of mean edge values into two in an iterative binary manner, till the above condition is satisfied. In our video datasets, we derive 7 frame complexity categories with 1 being the simplest and 7 the most complex. To speedup the online estimation of the accuracy on any candidate approximation branch, we create the offline accuracy profile $A_{B, F}$ given any frame complexity categories $F$ and any ABs $B$, after the 7 frame complexity categories are derived.

\noindent \textbf{Per-AB Contention-Aware Latency Profile}. 
\name is able to select ABs at runtime in the face of resource contention. Therefore, we perform offline profiling of the inference latency of each AB under different levels of contention. To study the resource contention, we develop our synthetic contention generator (CG) with tunable ``contention levels'' to simulate resource contention and help our \name profile and learn to react under such scenarios in real-life. Specifically, we run each AB in \name with the CG in varying contention levels to collect its contention-aware latency profile. To reduce the profiling cost, we quantize the contention to 10 levels for GPU and 20 levels for CPU and memory and then create the offline latency profile $L_{B, C}$ for each AB $B$ under each contention level $C$. Note that contention increases latency of the DNN but does not affect its accuracy. Thus, offline profiling for accuracy and latency can be done independently and parallelly and profiling overhead can be reduced.

\noindent \textbf{Switching Overhead Profile}. 
Since we find that the overhead of switching between some pairs of ABs is non-negligible, we profile the overhead of switching latency between any pairs of approximation branches offline. This cost is used in our optimization calculation to select the best AB. 

\subsection{Scheduler}
\label{subsec:pareto}

The main job of the scheduler in \name is to select an AB to execute. The scheduler accepts user requirement on either the minimum accuracy, the maximum latency per frame. The scheduler requests from the FCE a runtime accuracy profile ($B$ is the variable for the AB and $\hat{F}$ is the frame category of the input video frame) $A_{B, \hat{F}} \forall B$. It then requests from the RCE a runtime latency profile ($\hat{C}$ is the current contention level) $L_{B, \hat{C}} \forall B$. Given a target accuracy or latency requirement, we can easily select the AB to use from drawing the Pareto frontier for the current $(\hat{F}, \hat{C})$. If no Pareto frontier point satisfies the user requirement, \name picks the AB that achieves metric value closest to the user requirement. If the user does not set any requirement, \name sets a latency requirement to the frame interval of the incoming video stream. One subtlety arises due to the cost of switching from one AB to another. This cost has to be considered by the scheduler to avoid too frequent switches without benefit to outweigh the cost. 

To rigorously formulate the problem, we denote the set of ABs as $\mathcal{B}=\{B_1, ... B_N\}$ and the optimal AB the scheduler has to determine as $B_{opt}$. We denote the accuracy of branch $B$ on a video frame with frame complexity $F$ as $A_{B,F}$ , the estimated latency of branch $B$ under contention level $C$ as $L_{B,C}$, the one-time switch latency from branch $B_p$ to $B_{opt}$ as $L_{B_p \rightarrow B_{opt}}$, and the expected time window over which this AB can be used as $W$ (in the unit of frames). For $W$, we use the average number of frames for which the latest ABs can stay unchanged and this term introduces hysteresis to the system so that the AB does not switch back and forth frequently. The constant system overhead per frame (due to SCD, FCC, and resizing the frame) is $L_0$. Thus, the optimal branch $B_{opt}$, given the latency requirement $L_\tau$, is:

\begin{equation}
 B_{opt} = \argmax_{B \in \mathcal{B}} A_{B,F},~s.t.~L_{B,C}+\frac{1}{W}L_{B_p \rightarrow B}+L_0 \leq L_\tau
 \label{eq:opt-latency-constraint}
\end{equation}

When the accuracy requirement $A_\tau$ is given, 

\begin{equation}
  B_{opt} = \argmin_{B \in \mathcal{B}} [L_{B,C}+\frac{1}{W}L_{B_p \rightarrow B}+L_0], s.t.~A_{B,F} \geq A_\tau
  \label{eq:opt-latency}
\end{equation}

\section{Evaluation}
\label{sec_evaluation}

\subsection{Evaluation Platforms} \label{sec_platform}

We evaluate \name by running it on an NVIDIA Jetson TX2~\cite{tx2}, which includes 256 NVIDIA Pascal CUDA cores, a dual-core Denver CPU, a quad-core ARM CPU on a 8GB unified memory~\cite{unifor-mem} between CPU and GPU. The specification of this board is close to what is available in today's high-end smart phones such as Samsung Galaxy S20 and Apple iPhone 12. We train the approximation-enabled DNN on a server with NVIDIA Tesla K40c GPU with 12GB dedicated memory and an octa-core Intel i7-2600 CPU with 24GB RAM. For both the embedded device and the training server, we install Ubuntu 16.04 and TensorFlow v1.14.

\subsection{Datasets, Task, and Metrics}
\subsubsection{ImageNet VID dataset}

We evaluate \name on the video object classification task using ILSVRC 2015 VID dataset~\cite{ILSVRC2015_VID}. Although the dataset is initially used for object detection, we convert the dataset so that the task is to classify the frame into one of the ground truth object categories. If multiple objects exist, the classification is considered correct if matched with any one of the ground truth classes and this rule applies to both \name and baselines. According to our analysis, 89\% of the video frames are single-object-class frames and thus the accuracy is still meaningful under such conversion.

For the purpose of training, ILSVRC 2015 VID training set contains too many redundant video frames, leading to an over-fitting issue. To alleviate this problem, we follow the best practice in~\cite{kang2017t} such that the VID training dataset is sub-sampled every 180 frames and the resulting subset is mixed with ILSVRC 2014 detection (DET) training dataset to construct a new dataset with DET:VID=2:1. We use 90\% of this video dataset to train \name's DNN model and keep aside another 10\% as validation set to fine-tune \name (offline profiling). To evaluate \name's system performance, we use ILSVRC 2015 VID validation set -- we refer to this as the ``test set'' throughout the paper. 

\subsubsection{ImageNet IMG dataset}
We also use ILSVRC 2012 image classification dataset~\cite{deng2009imagenet} to evaluate the accuracy-latency trade-off of our single DNN. We use 10\% of the ILSVRC training set as our training set, first 50\% of the validation set as our validation set to fine-tune \name, and the remaining 50\% of the validation set as our test set. The choices made for training-validation-test in both the datasets follows common practice and there is no overlap between the three. \textbf{Throughout the evaluation, we use ImageNet VID dataset by default, unless we explicitly mention the use of the ImageNet IMG dataset.}

\subsubsection{Metrics}
We use latency and top-5 accuracy as the two metrics. The latency includes the overheads of the respective solutions, including the switching overhead, the execution time of FCE, RCE and scheduler.

\subsection{Baselines}\label{sec:baseline}

We start with the evaluation on the static models without the ability to adapt. This is because we want to reveal the relative accuracy-latency trade-offs in the traditional settings, compared to the single approximation branches in \name. The baselines for this static experiment include model variants, which are designed for different accuracy and latency goals, \ie ResNet~\cite{resnet} MobileNets~\cite{howard2017mobilenets} and  MSDNets~\cite{huang2017multi} for which we use 5 execution branches in the single model that can provide different accuracy and latency goals. We use the ILSVRC IMG dataset to evaluate these static models, since this dataset is larger and with more classes.

We then proceed with the evaluation on the streaming videos, under varying resource contention. This brings two additional aspects for evaluation -- (1) how the video frames are being processed in a timely and streaming manner as frames in this case cannot be batched like images, and (2) how the technique can meet the latency budget in the presence of resource contention from other application that can raise the processing latency. 
The baselines we use are: MCDNN~\cite{mcdnn} as a representative of the multi-model approach, and MSDNets, a representative of the single-model approach (with multiple execution branches). 
We also compare the switching overhead of our single-model design with the multi-capacity models in NestDNN~\cite{fang2018nestdnn}.
Unfortunately, we were not able to use BranchyNet~\cite{teerapittayanon2016branchynet}, because their DNN is not designed for large images in the ImageNet dataset. BranchyNet was evaluated on MNIST and CIFAR datasets in thir paper and it does not provide any guidance on the parameter settings for training and makes it impossible to use on different datasets. 

The details of each baseline are as follows,

\noindent \textbf {ResNet}: ResNet is the base DNN architecture of many state-of-the-art image and video object classification tasks, with superior accuracy to other architectures. While it was originally meant for server-class platforms, as resources on mobile devices increase, ResNet is also being used on such devices~\cite{lu2017modeling, zhang2018shufflenet, wang2018pelee}. We use ResNet of 18 layers (ResNet-18) and of 34 layers (ResNet-34) as base models. We modify the last FC layer to classify into 30 labels in the VID dataset and fine-tune the whole model. ResNet-34 plays a role as the reference providing the upper bound of the target accuracy. ResNet architectures with more than 34 layers (\cite{resnet} has considered up to 152 layers) become impractical as they are too slow to run on the resource-constrained mobile devices and their memory consumption is too large for the memory on the board.  

\noindent \textbf {MobileNets}: This refers to 20 model variants (trained by the original authors) specifically designed for mobile devices ($\alpha=1,0.75,0.5,0.35, shape=224,192,160,128,96$).

\noindent \textbf {MSDNets}: This refers to the 5 static execution branches to meet the different latency budgets in their anytime evaluation scenario. We have enhanced MSDNets with a scheduler to dynamically choose the static branches for dynamic runtime conditions. The former is compared with static models in the IMG dataset and the latter is compared with adaptive systems in the VID dataset. For the sake of simplicity, we reuse the same term (MSDNets) to refer to both.

\noindent \textbf {NestDNN}: This solution provides multi-capacity models with ResNet-34 architecture by varying the number of filters in each convolutional layer. We compose 9 descendant models, where (1) the seed model, or the smallest model, reduces the number of filters of all convolutional layers uniformly by 50\%, (2) the largest descendant model is exactly ResNet of 34 layers, and (3) the other descendant models reduce the number of filters of all convolutional layers uniformly by a ratio equally distributed between 50\% and 100\%. We only compare the switching overhead of NestDNN inside its descendant models with \name because NestDNN is not open-sourced and the paper does not provide enough details about the training process or the architecture.

\noindent \textbf {MCDNN}: We change the base model in MCDNN from VGG to the more recent ResNet for a fairer comparison. This system chooses between MCDNN-18 and MCDNN-34 depending on the accuracy requirement. MCDNN-18 uses two models: a specialized ResNet-18 followed by the generic ResNet-18. The specialized ResNet-18 is the same as the ResNet-18 except the last layer, which is modified to classify the most frequent $N$ classes only. This is MCDNN's key novelty that most inputs belong to the top $N$ classes, which can be handled by a reduced-complexity DNN. If the top-1 prediction label of the specialized model in MCDNN is not among the top $N$ frequent classes, then the generic model processes the input again and outputs its final predictions. Otherwise, MCDNN uses the top-5 prediction labels of the specialized model as its final predictions. We set $N=20$ that covers 80\% of training video frames in the VID dataset.

\subsection{Typical Usage Scenarios}
\label{sec:scenarios}

We use a few usage scenarios to compare the protocols, although \name can support finer-grained user requirements in latency or accuracy. 

\begin{itemize}
    \item \textbf{High accuracy, High latency (HH)} refers to the scenario where \name has less than 10\% (relative) accuracy loss from ResNet-34, our most accurate single model baseline. Accordingly, the runtime latency is also high to achieve such accuracy.
    \item \textbf{Medium accuracy, Medium latency (MM)} has an accuracy loss less than 20\% from our base model ResNet-34.
    \item \textbf{Low accuracy, Low latency (LL)} can tolerate an accuracy loss of up to 30\% with a speed up in its inferencing.
    \item \textbf{Real time (RT)} scenario, by default, means the processing pipeline should keep up with 30 fps speed, \ie maximum 33.33 ms latency. This is selected if no requirement is specified. 
\end{itemize}

\begin{figure*}[b]
  \centering
  \begin{minipage}[t]{0.49\linewidth} 
    \includegraphics[width=0.99\textwidth]{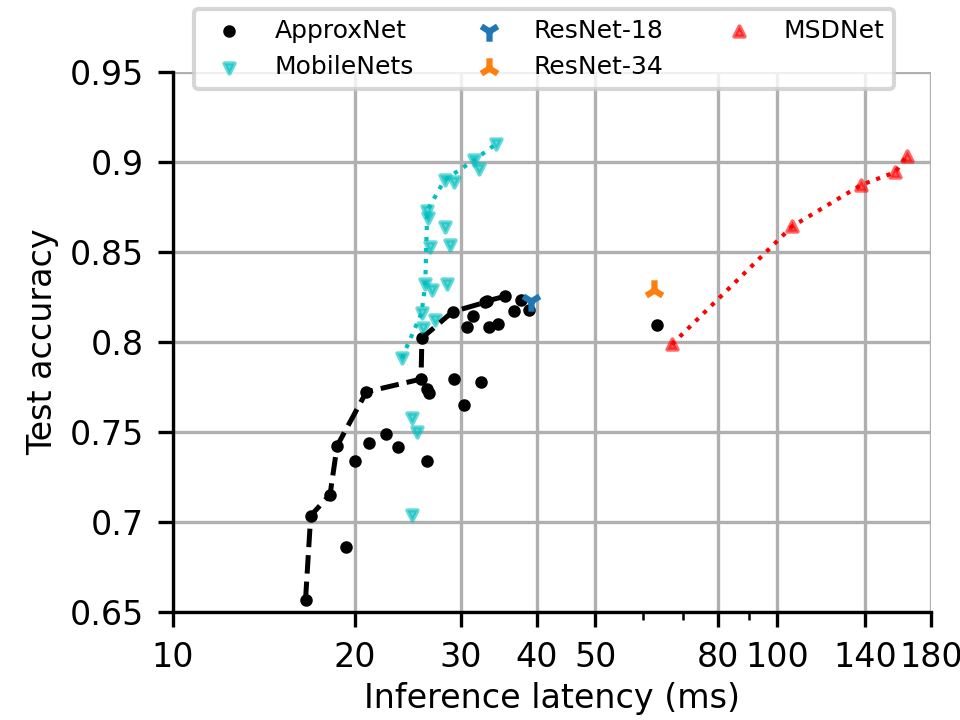}
    \caption{Pareto frontier for test accuracy and inference latency on the ImageNet IMG dataset for ApproxNet compared to ResNet and MobileNets, the latter being specialized for mobile devices.}
    \label{trade_off_img_dataset}
  \end{minipage}
  \hfill
  \begin{minipage}[t]{0.49\linewidth} 
    \centering
    \includegraphics[width=0.99\textwidth]{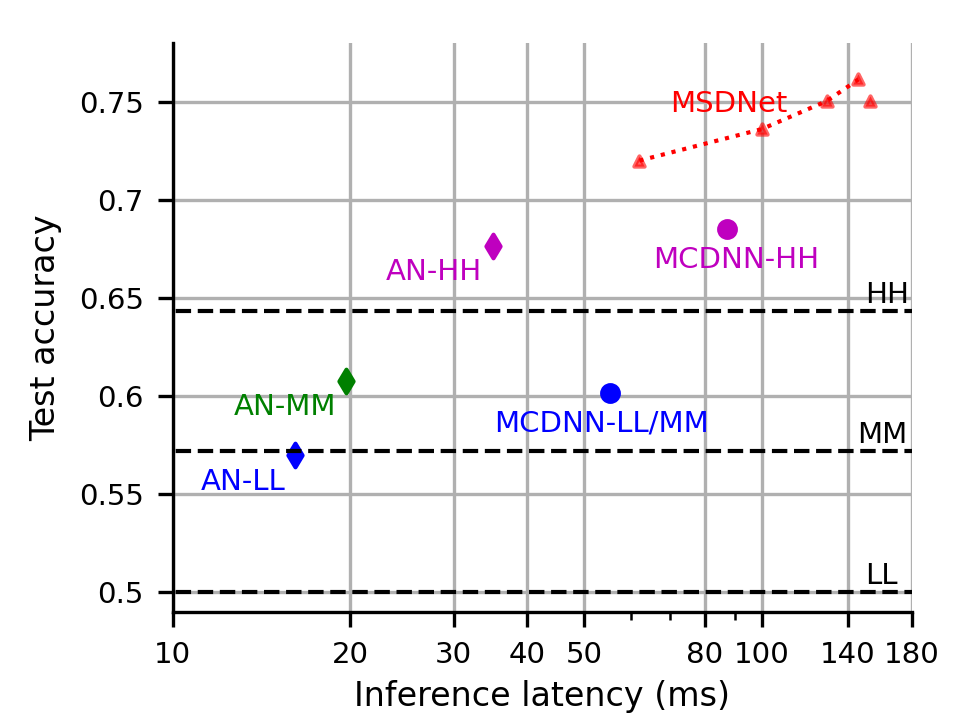}
    \caption{Comparison of system performance in typical usage scenarios. ApproxNet is able to meet the accuracy requirement for all three scenarios. User requirements are shown in dashed lines.}
    \label{trade_off_system}
  \end{minipage}
\end{figure*}

\begin{table}[t]
  \centering
  \caption{Averaged accuracy and latency performance of ABs on the Pareto frontier in ApproxNet and those of the baselines on validation set of the VID dataset. Note that the accuracy on the validation dataset can be higher due to its similarity with the training dataset. Thus the validation accuracy is only used to construct the look up table in \name and baselines and does not reflect the true performance.}
  \label{table:LUT_VID}
  \begin{minipage}[]{0.99\linewidth}
    \centering
    \centerline{(a) Averaged accuracy and latency performance in ApproxNet.}
    \begin{tabular}{lcccc} 
      \hline
      Usage Scenario (rate) & Shape      &   Layers & Latency  & Accuracy \\
      \hline
      {\bf HH} (32 fps)     & 128x128x3  &   24     & 31.42 ms & 82.12\% \\
                            & 160x160x3  &   20     & 31.33 ms & 80.81\% \\
                            & 128x128x3  &   20     & 27.95 ms & 79.35\% \\
                            & 112x112x3  &   20     & 26.84 ms & 78.28\% \\
      {\bf MM} (56 fps)     & 128x128x3  &   12     & 17.97 ms & 70.23\% \\
                            & 112x112x3  &   12     & 17.70 ms & 68.53\% \\
                            & 96x96x3    &   12     & 16.78 ms & 67.98\% \\ 
      {\bf LL} (62 fps)     & 80x80x3    &   12     & 16.14 ms & 66.39\% \\
      \hline         
    \end{tabular}
  \end{minipage}
  \begin{minipage}[]{0.99\linewidth}
    \centering
    \centerline{(b) Lookup table in MCDNN's scheduler.}
    \begin{tabular}{lcccc} 
      \hline
      Scenario (rate)      & Shape      &   Layers & Latency  & Accuracy \\
      \hline
      {\bf HH} (11 fps)    & 224x224x3  &   34     & 88.11 ms & 77.71\%  \\
      {\bf MM/LL} (17 fps) & 224x224x3  &   18     & 57.83 ms & 71.40\%  \\
      \hline         
    \end{tabular}
  \end{minipage}
  \begin{minipage}[]{0.99\linewidth}
    \centering
    \centerline{(c) Lookup table in MSDNet's scheduler.}
    \begin{tabular}{lcccc} 
      \hline
      Scenario (rate)      & Shape      &   Layers & Latency  & Accuracy \\
      \hline
      {\bf HH} (5.2 fps)   & 224x224x3  &   191    & 153 ms   & 96.79\%  \\
      {\bf MM/LL} (16 fps) & 224x224x3  &   63     & 62 ms    & 95.98\%  \\
      \hline         
    \end{tabular}
  \end{minipage}
  \begin{minipage}[]{0.99\linewidth}
    \centering
    \centerline{(d) Reference performance of single model variants or execution branches.}
    \begin{tabular}{ccccl} 
      \hline
      Model name (rate)    & Shape      &   Layers & Latency  & Accuracy \\
      \hline
      ResNet-34 (16 fps)   & 224x224x3  &   34     & 64.44 ms & 85.86\%  \\ 
      ResNet-18 (22 fps)   & 224x224x3  &   18     & 45.22 ms & 84.59\%  \\
      MSDNet-branch5 (5.2 fps) & 224x224x3 & 191   & 153 ms & 96.79\%  \\
      MSDNet-branch4 (5.6 fps) & 224x224x3 & 180   & 146 ms & 96.55\%  \\
      MSDNet-branch3 (7.8 fps) & 224x224x3 & 154   & 129 ms & 96.70\%  \\
      MSDNet-branch2 (10 fps) & 224x224x3 & 115    & 100 ms & 96.89\%  \\
      MSDNet-branch1 (16 fps) & 224x224x3 & 63     & 62 ms  & 95.98\%  \\
      \hline         
    \end{tabular}
  \end{minipage}
\end{table}

\begin{figure*}[t]
  \centering
  \includegraphics[width=0.9\linewidth]{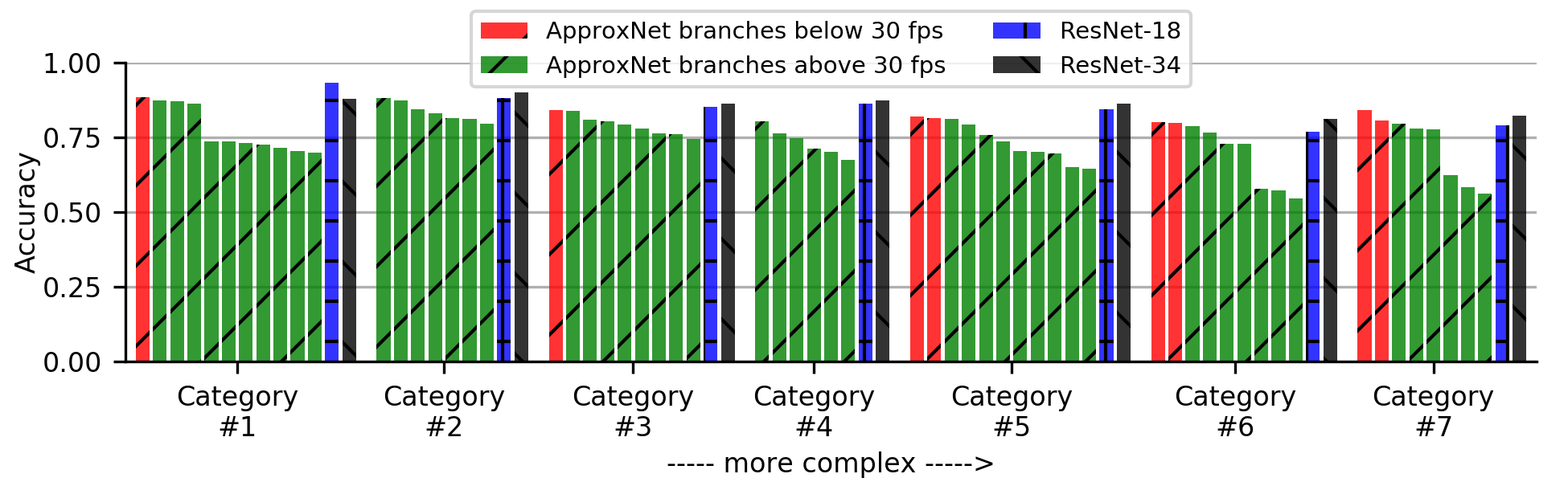}
  \caption{Content-specific accuracy of Pareto frontier branches. Branches that fulfill real-time processing (30 fps) requirement are labeled in green. Note that both ResNet-18 and ResNet-34 models, though with the higher accuracy, cannot meet the 30 fps latency requirement.}
  \label{per_category_trade_off}
\end{figure*}
                                                                                                     
\subsection{Accuracy-latency Trade-off of the Static Models}

We first evaluate \name on ILSVRC IMG dataset on the accuracy-latency trade-off of each AB in our single DNN as shown in Figure~\ref{trade_off_img_dataset}. Our AB with higher latency (satisfying 30 fps speed though higher) has close accuracy to ResNet-18 and ResNet-34 but much lower latency than ResNet-34. Meanwhile, our AB with reduced latency (25 ms to 30 ms) has close accuracy to MobileNets. And finally, our AB is superior to all baselines in achieving extremely low latency (< 20 ms). However, the single execution branches in MSDNet are much slower than \name and other baselines. The latency ranges from 62 ms to 153 ms, which cannot meet the real-time processing requirement and will be even worse in face of the resource contention. MobileNets can keep up with the frame rate but it lacks the configurability in the latency dimensional that \name has. Although MobileNets does win on the IMG dataset at higher accuracy, it needs an ensemble of models (like MCDNN) when it comes to the video where content characteristics, user requirement, and runtime resource contention change.

\subsection{Adaptability to Changing User Requirements} \label{sec:adp_user_req}

We then, from now on, switch to the ILSVRC VID dataset and show how \name can meet different user requirements for accuracy and latency. We list the averaged accuracy and latency of Pareto frontier branches in Table~\ref{table:LUT_VID}(a), which can serve as a lookup table in the simplest scenario, \ie without considering frame complexity categories and resource contention. \name, provides content-aware approximation and thus keeps a lookup table for each frame complexity category, and to be responsive to resource contention, updates the latency in the lookup table based on observed contention. 

We perform our evaluation on the entire test set, but without the baseline protocols incurring any switching penalty. Figure~\ref{trade_off_system} compares the accuracy and latency performance between \name and baselines in three typical usage scenarios ``HH'', ``MM'', and ``LL'' (AN denotes \name). In this experiment, \name uses the content-aware lookup table for each frame complexity category and chooses the best AB at runtime to meet the user accuracy requirement. MCDNN and MSDNet use similar lookup tables (Table~\ref{table:LUT_VID}(b) and (c)) to select among model variants or execution branches to satisfy the user requirement. We can observe that ``AN-HH'' achieves the accuracy of 67.7\% at a latency of 35.0 ms, compared to ``MCDNN-HH'' that has an accuracy of 68.5\% at the latency of 87.4 ms. Thus, MCDNN-HH is 2.5X slower while achieving 1.1\% accuracy gain over \name. On the other hand, MSDNet is more accurate and slower than all \name's branches. The lightest branch and heaviest branch achieve 4.3\% and 7.3\% higher accuracy respectively, and incur 1.8X and 4.4X higher latency respectively. In ``LL'' and ``MM'' usage scenarios, MCDNN-LL/MM is 2.8-3.3X slower than \name, while gaining in accuracy 3\% or less. MSDNets, on the other hand, is running with much higher latency (62 ms to 146 ms) and higher accuracy (72.0\% to 76.2\%). 
Thus, compared to these baseline models, \name wins by providing lower latency, satisfying the real-time requirement, and flexibility in achieving various points in the (accuracy, latency) space. 

\subsection{Adaptability to Changing Content Characteristics \& User Requirements}\label{system_latency}

We now show how \name can adapt to changing content characteristics and user requirements within the same video stream. The video stream, typically at 30 fps, may contain content of various complexities and this can change quickly and arbitrarily. Our study with the FCC on the VID dataset has shown that in 97.3\% cases the frame complexity category of the video will change within every 100 frames. Thus, dynamically adjusting the AB with frame complexity category is beneficial to the end-to-end system. We see in Figure~\ref{per_category_trade_off} that \name with various ABs can satisfy different (accuracy, latency) requirements for each frame complexity category. According to user's accuracy or latency requirement, \name's scheduler picks the appropriate AB. The majority of the branches satisfy the real-time processing requirement of 30 fps and can also support high accuracy quite close to the ResNet-34.

\begin{figure*}[t]
  \centering
  \begin{minipage}[thb]{0.4\linewidth} 
    \includegraphics[width=1\textwidth]{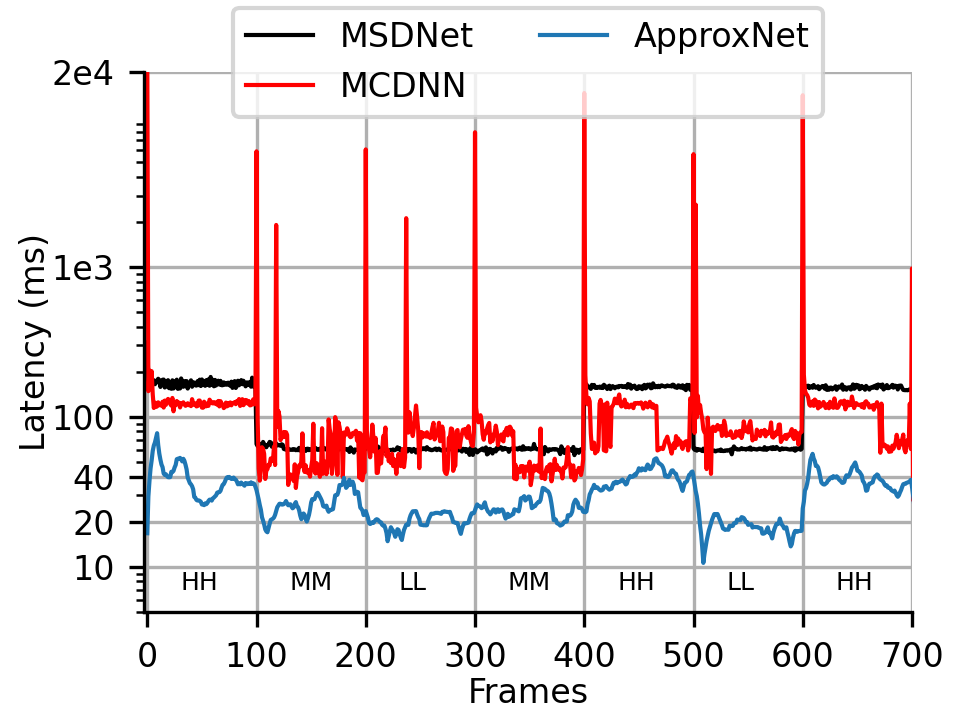}
    \caption{Latency performance comparison with changing user requirements throughout video stream.}
    \label{temporal_system}
  \end{minipage}
  \hfill
  \begin{minipage}[thb]{0.58\linewidth}
    \begin{minipage}[thb]{0.47\linewidth}
        \includegraphics[width=1\textwidth]{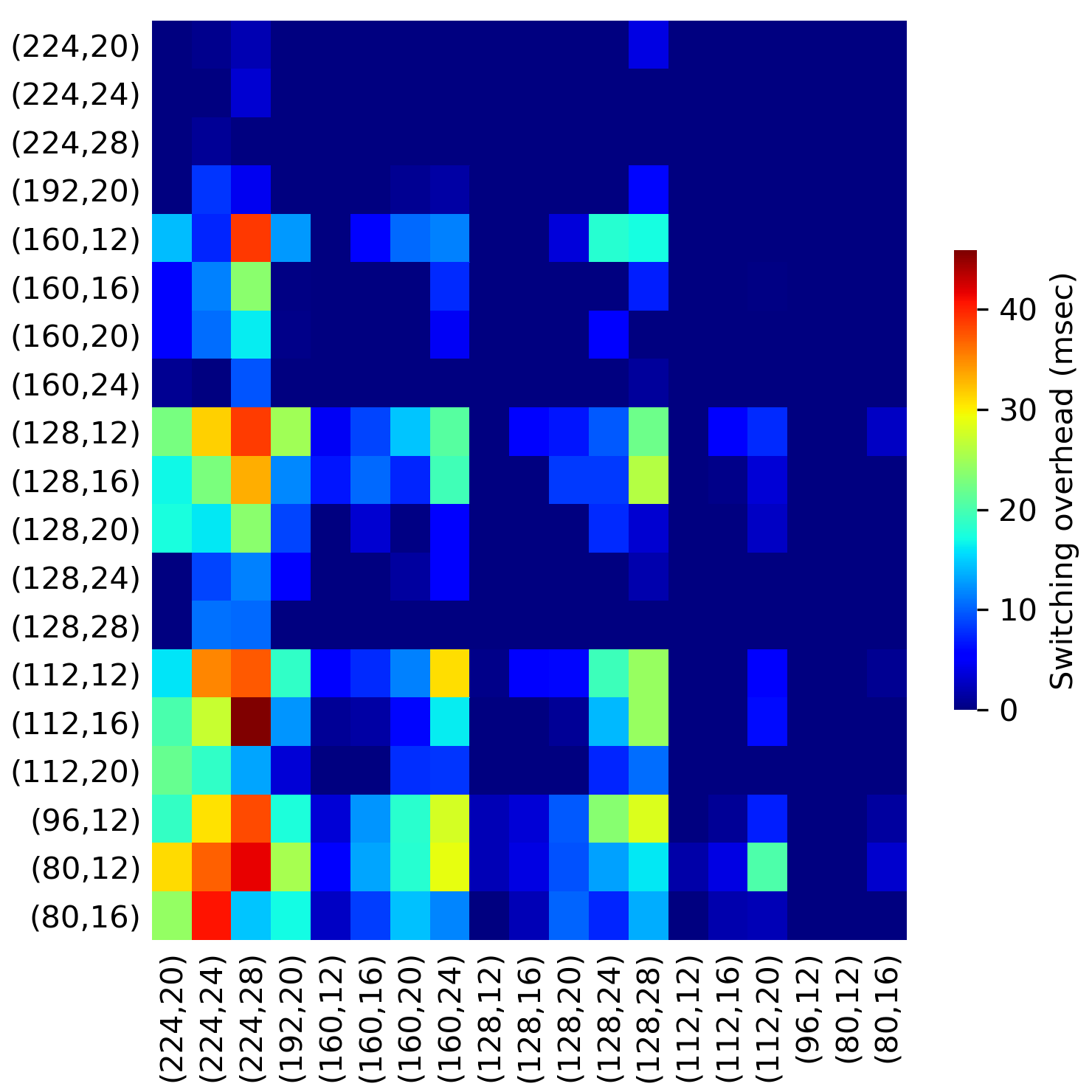}
        \centerline{(a) \name}
    \end{minipage} 
    \begin{minipage}[thb]{0.52\linewidth}
        \includegraphics[width=1\textwidth]{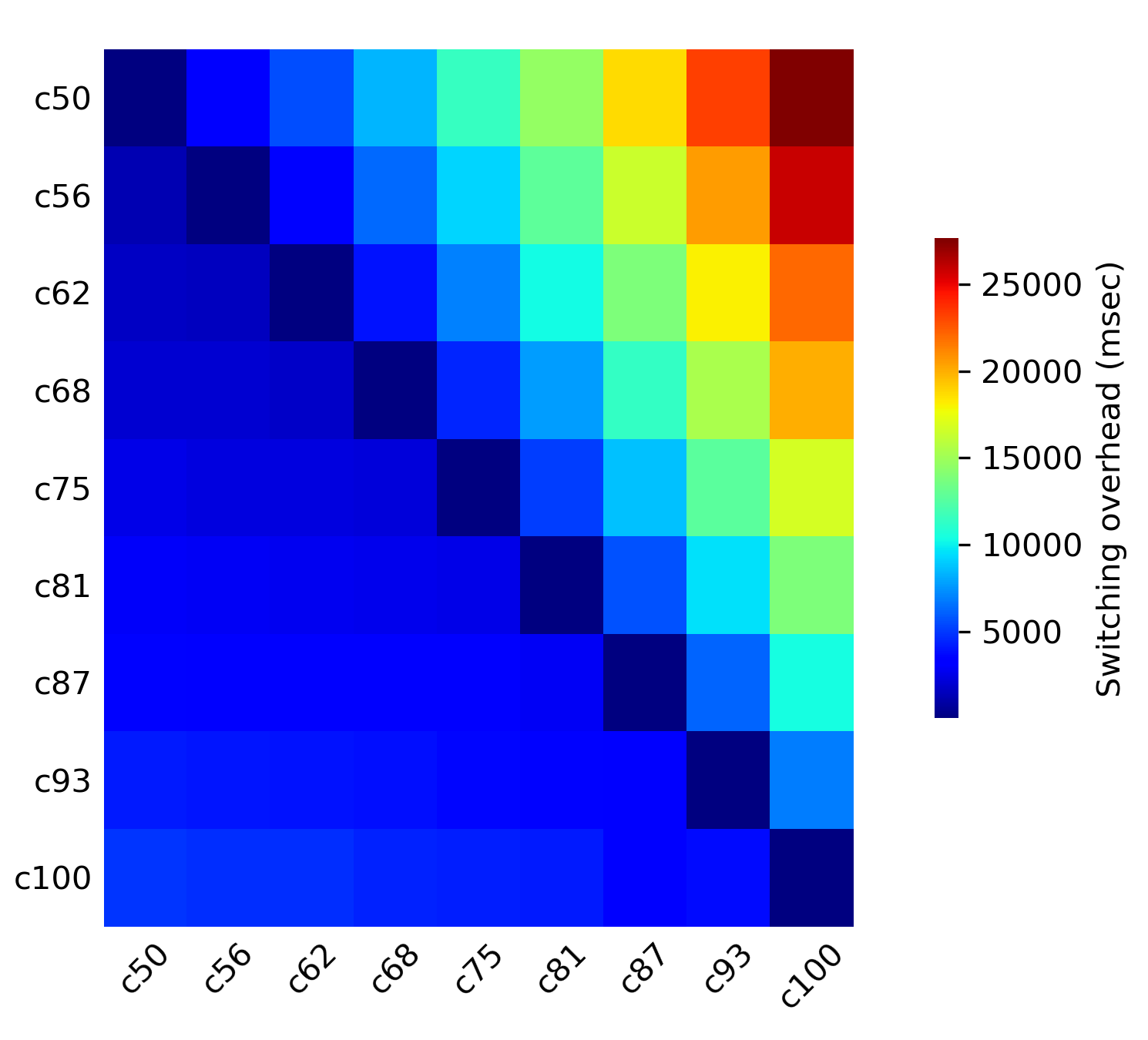}
        \centerline{(b) NestDNN}
    \end{minipage} 
    \caption{Transition latency overhead across (a) ABs in ApproxNet and (b) descendant models in NestDNN. ``from'' branch on Y-axis and ``to'' branch on X-axis. Inside brackets: (input shape, outport depth). Latency unit is millisecond.}
    \label{fig:transition}
  \end{minipage} 
\end{figure*}

In Figure~\ref{temporal_system}, we show how \name adapts for a particular representative video from the test dataset. Here, we assume the user requirement changes every 100 frames between ``HH'', ``MM'', and ``LL''. This is a synthetic setting to observe how models perform at the time of switching. We assume a uniformly distributed model selection among 20 model variants for MCDNN's scheduler (in~\cite{mcdnn}, the MCDNN catalog uses 68 model variants) while the embedded device can only cache two models in the RAM (more detailed memory results in Section~\ref{subsec:overhead}). In this case, MCDNN has a high probability to load a new model variant into RAM from Flash, whenever the user requirement changes. This results in a huge latency spike, typically from 5 to 20 seconds at each switch. It is notable that for some cases, there are also small spikes in MCDNN following the larger spikes because the generic model is invoked due to the specialized model's prediction of ``infrequent'' class. On the other hand, \name and MSDNets incur little overhead in switching between any two branches, because they are all available within the same single-model DNN. Similar to the results before, \name wins over MSDNets at lower latency to meet the real-time processing requirement even though its accuracy is slightly lower.

\begin{figure*}[t]
  \centering
  \begin{minipage}[t]{0.49\linewidth}
    \includegraphics[width=1\textwidth]{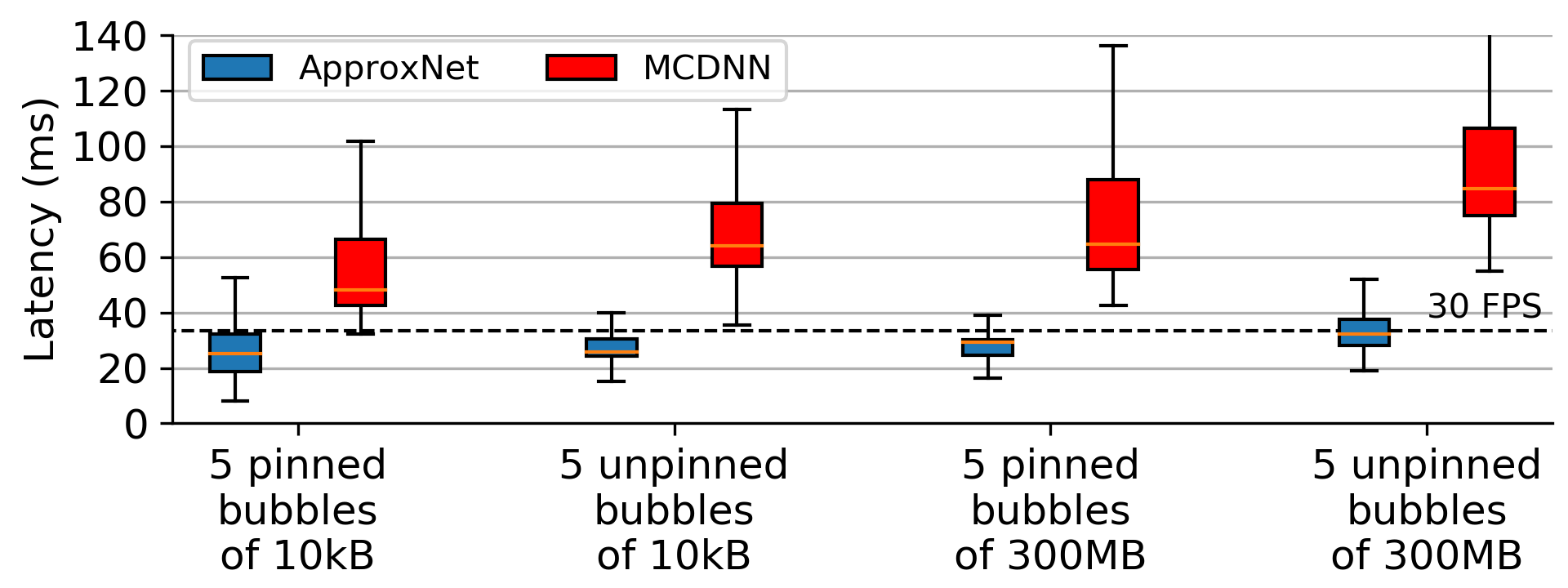}
    \centerline{(a) Inference latency (w/ CPU contention)}
  \end{minipage}
  \hfill
  \begin{minipage}[t]{0.49\linewidth}
    \centering
    \includegraphics[width=1\textwidth]{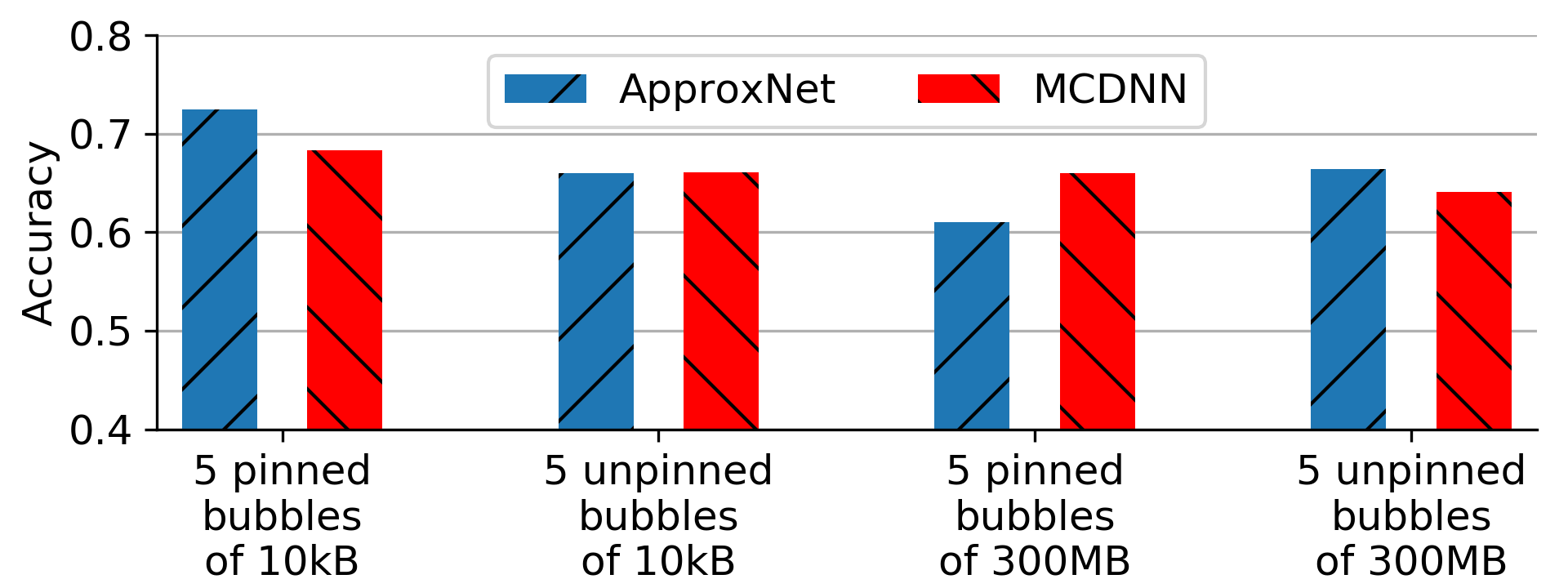}
    \centerline{(b) Accuracy (w/ CPU contention)}
  \end{minipage}
  \hfill
  \begin{minipage}[t]{0.49\linewidth}
    \centering
    \includegraphics[width=1\textwidth]{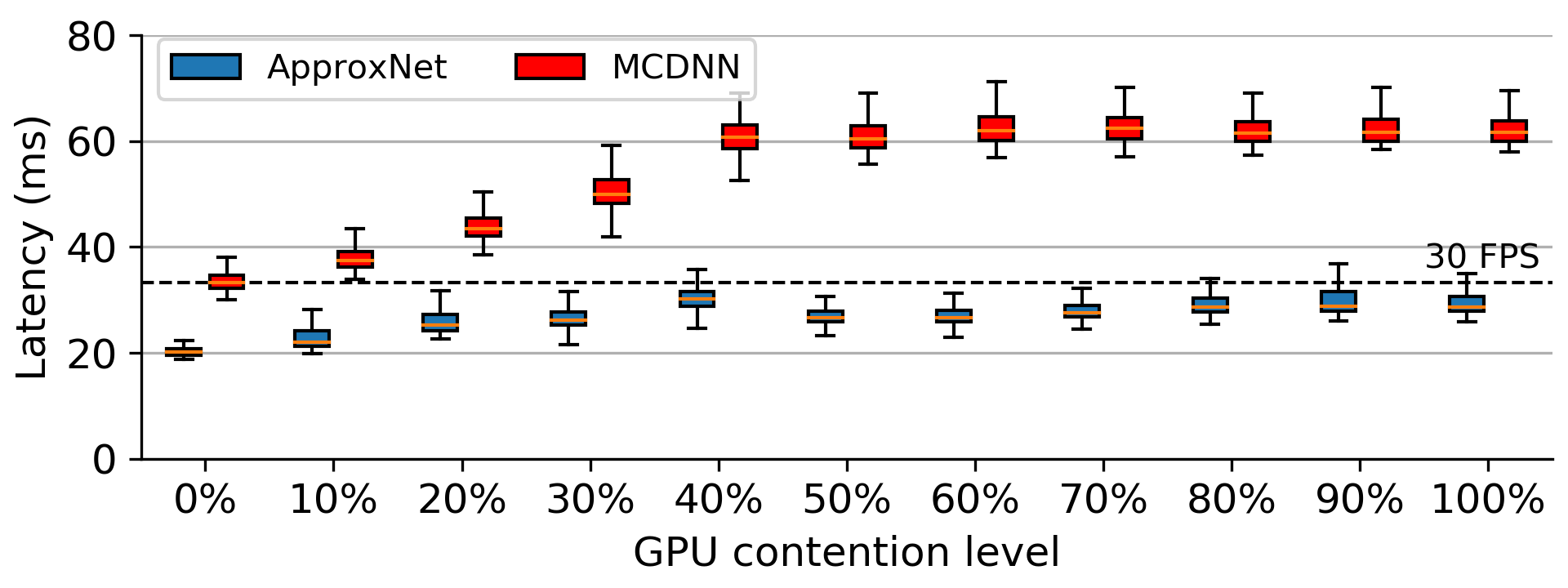}
    \centerline{(c) Inference latency (w/ GPU contention)}
  \end{minipage}
  \caption{Comparison of ApproxNet vs MCDNN under resource contention. (a) and (b) inference latency and accuracy on the whole test dataset. (c) inference latency on a test video.}
  \label{fig:system_contention}
\end{figure*}  

To see in further detail the behavior of \name, we profile the mean transition time of all Pareto frontier branches under no contention as shown in Figure~\ref{fig:transition} (a). Most of the transition overheads are extremely low, while only a few transitions are above 30 ms. Our optimization algorithm (Equations~\ref{eq:opt-latency-constraint} and~\ref{eq:opt-latency}) filters out such expensive transitions if they happen too frequently. In Figure~\ref{fig:transition} (b), we further show the transition time between the descendant models in NestDNN, which uses a multi-capacity model with varying number of filters in the convolutional layers. The notation c50 stands for the descendant model with 50\% capacity of the largest variant, and so on. We can observe that the lowest switching cost of NestDNN is still more than an order of magnitude higher than the highest switching cost of \name. The highest switching cost of NestDNN compared to the highest of \name is more than three orders of magnitude---the highest cost is important when trying to guarantee latencies with the worst-case latency spikes. We can observe that the transition overhead can be up to 25 seconds from the smallest model to the largest model, and is generally proportional to the amount of data that is loaded into the memory. This is because NestDNN only keeps a single model, the one in use in memory and loads/unloads all others when switching. In summary, the benefit of \name comes from the fact that (1) it can accommodate multiple (accuracy, latency) points within one model through its two approximation knobs while MCDNN has to switch between model variants, and (2) switching between ABs does not load large amount of data and computational graphs. 

\subsection{Adaptability to Resource Contention}
\label{resource_contention}

We evaluate in Figure~\ref{fig:system_contention}, the ability of \name to adapt to resource contention on the device, both CPU and GPU contention. First, we evaluate this ability by running a \textit{bubble application}~\cite{mars2011bubble, xu2018pythia} on the CPU that creates stress of different magnitudes on the (shared) memory subsystem while the video analytics DNN is running on the GPU. We generate bubbles, of two different memory sizes 10 KB  (low contention) and 300 MB (high contention). The bubbles can be ``unpinned'' meaning they can run on any of the cores or they can be ``pinned'' in which case they run on a total of 5 CPU cores leaving the 6th one for dedicated use by the video analytics application. The unpinned configuration causes higher contention. We introduce contention in phases---low pinned, low unpinned, high pinned, high unpinned. 

As shown in Figure~\ref{fig:system_contention}(a), MCDNN with its fastest model variant MCDNN-18, runs between 40ms and 100 ms depending on the contention level and has no adaptation. For \name, on the other hand, our mean latency under low contention (10 KB, pinned) is 25.66 ms, and it increases a little to 34.23 ms when the contention becomes high (300 MB, unpinned). We also show the accuracy comparison in Figure~\ref{fig:system_contention}(b), where we are slightly better than MCDNN under low contention and high contention (2\% to 4\%) but slightly worse (within 4\%) for intermediate contention (300 MB, pinned). 

To further evaluate \name with regard to GPU contention, we run a synthetic matrix manipulation application concurrently with \name. The contention level is varied in a controlled manner through the synthetic application, from 0\% to 100\% in steps of 10\%, where the control is the size of the matrix, and equivalently, the number of GPU threads dedicated to the synthetic application. The contention value is the GPU utilization when the synthetic application runs alone as measured through \texttt{tegrastats}. For baseline, we use the MCDNN-18 model again since among the MCDNN ensemble, it comes closest to the video frame rate (33.3 ms latency). As shown in Figure~\ref{fig:system_contention}(c), without the ability to sense the GPU contention and react to it, the latency of MCDNN increases by 85.6\% and is far beyond the real-time latency threshold. The latency of \name also increases with gradually increasing contention, 20.3 ms at no contention to 30.77 ms at 30\% contention. However, when we further raise the contention level to 50\% or above, \name's scheduler senses the contention and switches to a lighter-weight approximation branch such that the latency remains  within 33.3 ms. The accuracy of MCDNN and \name were identical for this sample execution. Thus, this experiment bears out the claim that \name can respond to contention gracefully by recreating the Pareto curve for the current contention level and picking the appropriate AB.

\begin{figure*}[t]
  \begin{minipage}[thb]{0.44\linewidth}
      \begin{minipage}[thb]{1\linewidth}
        \centering
        \includegraphics[width=1\textwidth]{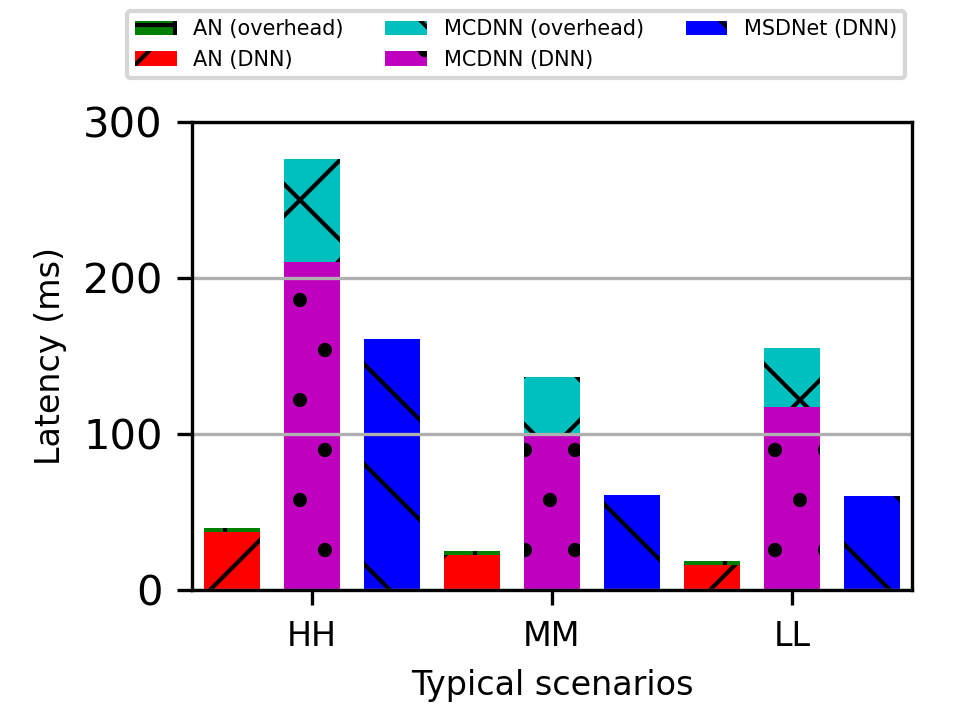}
        \caption{System overhead in ApproxNet and MCDNN.}
        \label{fig:timing_share}
      \end{minipage}
      \begin{minipage}[thb]{1\linewidth}
        \centering
        \includegraphics[width=1\textwidth]{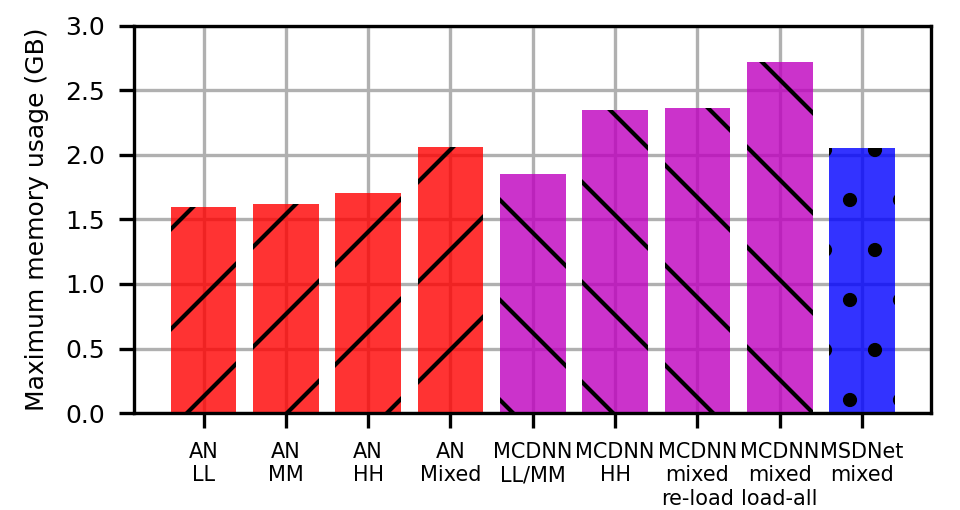}
        \caption{Memory consumption of solutions in different usage scenarios (unit of GB).}
        \label{fig:memory-result}
      \end{minipage}
  \end{minipage}
  \hfill
  \begin{minipage}[thb]{0.53\linewidth}
    \centering
    \includegraphics[width=1\textwidth]{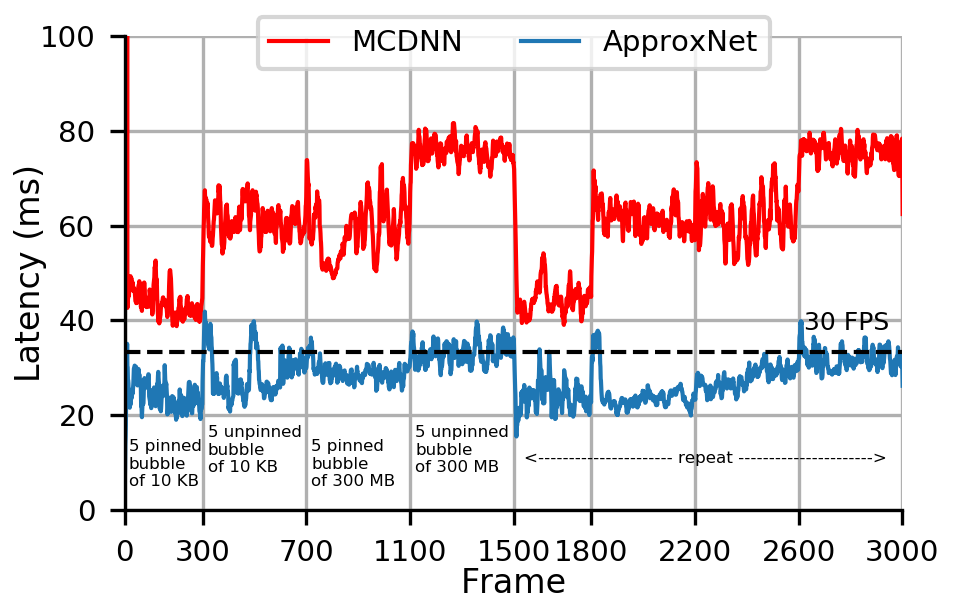}
    \centerline{(a) Inference latency}
    \includegraphics[width=1\textwidth]{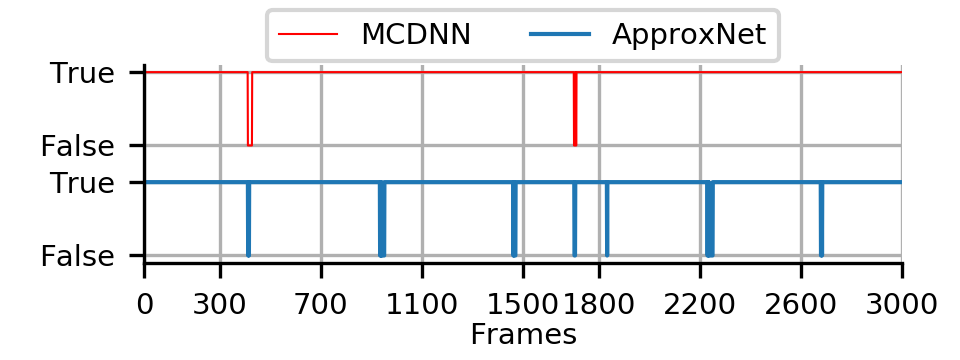}
    \centerline{(b) Accuracy}
    \caption{Case study: performance comparison of ApproxNet vs MCDNN under resource contention for a Youtube video.}
    \label{fig:system_contention_case_study}
  \end{minipage}
\end{figure*}

\subsection{Solution Overheads} \label{subsec:overhead}

With the same experiment as in Section~\ref{system_latency}, we compare the overheads of \name, MCDNN, and MSDNets in Figure~\ref{fig:timing_share}. For \name, we measure the overhead of all the steps outside of the core DNN, \ie frame resizing, FCE, RCE, and scheduler. For MCDNN, the dominant overhead is the model switching and loading. The model switching overhead of MCDNN is measured at each switching point and averaged across all frames in each scenario. We see that \name, including overheads, is $7.0X$ to $8.2X$ faster than MCDNN and $2.4X$ to $4.1X$ faster than MSDNets. Further, we can observe that in ``MM'' and ``LL'' scenarios, \name's averaged latency is less than 30 ms and thus \name can achieve real-time processing of 30 fps videos. As mentioned before, MCDNN may be forced to reload the appropriate models whenever the user requirement changes. So, in the best case for MCDNN the requirement never changes or it has all its models cached in RAM. \name is still $5.1X$ to $6.3X$ faster.

Figure~\ref{fig:memory-result} compares the peak memory consumption of \name and MCDNN in typical usage scenarios. \name-mixed, MCDNN-mixed, and MSDNet-mixed are the cases where the experiment cycles through the three usage scenarios. We test MCDNN-mixed with two model caching strategies: (1) the model variants are loaded from Flash when they get triggered (named ``re-load''), simulating the minimum RAM usage (2) the model variants are all loaded into the RAM at the beginning (named ``load-all''), assuming the RAM is large enough. We see that \name in going from ``LL'' to ``HH'' requirement consumes 1.6 GB to 1.7 GB memory and is lower than MCDNN (1.9 GB and 2.4 GB). MCDNN's cascade DNN design (specialized model followed by generic model) is the root cause that it consumes about 15\% more memory than our model even though they only keep one model variant in the RAM and it consumes 32\% more memory if it loads two. For the mixed scenario, we can set an upper bound on the \name memory consumption---it never exceeds 2.1 GB no matter how we switch among ABs at runtime, an important property for proving operational correctness in mobile or embedded environments. Further, \name, with tens of ABs available, offers more choices than MCDNN and MSDNet, and MCDNN cannot accommodate more than two models in the available RAM. 

Storage is a lesser concern but it does affect the pushing out of updated models from the server to the mobile device, a common use case. \name's storage cost is only 88.8 MB while MCDNN with 2 models takes 260 MB and MSDNet with 5 execution branches takes 177 MB. A primary reason is the duplication in MCDNN of the specialized and the generic models which have identical architecture except for the last FC layer. Thus, \name is well suited to the mobile usage scenario due to its low RAM and storage usage.

\subsection{Ablation Study with FCE} \label{sec_ablation_study_FCE}

\begin{figure}[t]
  \centering
  \includegraphics[width=0.5\linewidth]{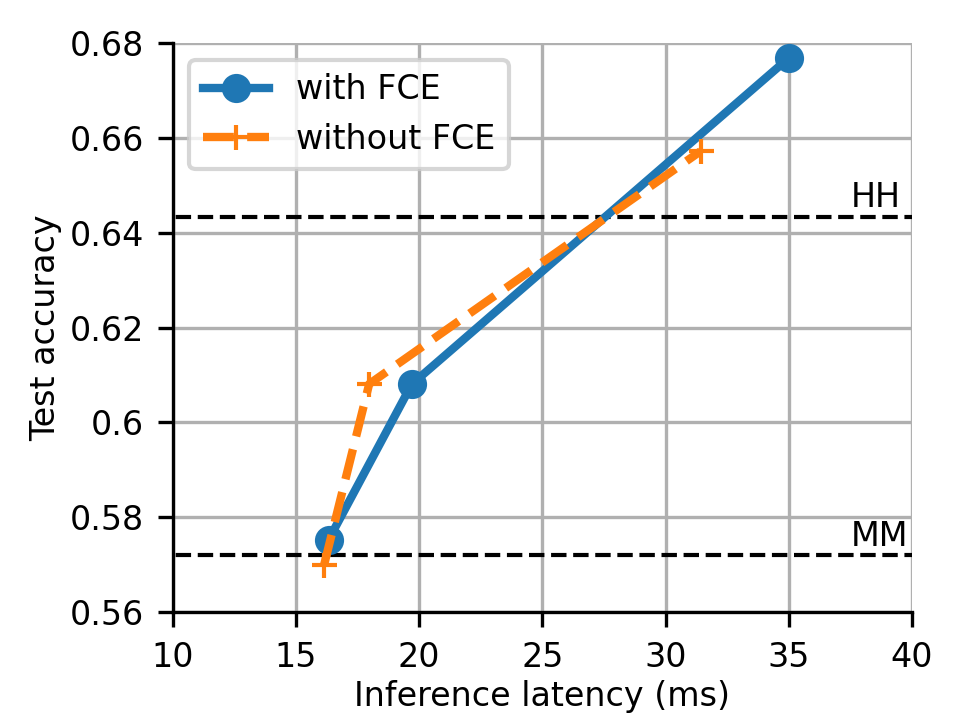}
  \caption{Comparison of system performance in typical usage scenarios between \name with FCE and \name without FCE.}
  \label{fig:ablation_FCE}
\end{figure}

To study the necessity of FCE, we conduct an ablation study of \name with a content agnostic scheduler. Different from the content-aware scheduler, this scheduler picks the same AB for all video frames regardless of the content (the contention level is held unchanged). The test dataset obviously has variations in the content characteristics. With the same experiment as in Section~\ref{sec:adp_user_req}, we compare the accuracy and latency performance between \name with FCE and that without FCE in three typical usage scenarios ``HH'', ``MM'', and ``LL''. Figure~\ref{fig:ablation_FCE} shows that \name with FCE is able to improve the accuracy by 2.0\% under a ``HH'' scenario with an additional 3.57 ms latency cost. \name with FCE under ``MM'' and ``LL'' scenarios is either slightly slower or more accurate than that without FCE. To summarize, \name's FCE is more beneficial when the accuracy goal is higher, and however the latency budget is also higher.

\subsection{Case Study with YouTube Video} \label{sec_case_study}

As a case study, we evaluate \name on a randomly picked YouTube video~\cite{YoutubeVideoLink}, to see how it adapts to different resource contention scenarios at runtime (Figure~\ref{fig:system_contention_case_study}). The video is a car racing match with changing scenes and objects, and thus, we want to evaluate the object classification performance. The interested reader may see a demo of \name and MCDNN on this and other videos at \url{https://approxnet.github.io/}. Similar to the control setup in Section~\ref{resource_contention}, we test \name and MCDNN for four different contention levels. Each phase is 300-400 frames and the latency requirement is 33 ms to keep up with the 30 fps video. We see \name adapts to the resource contention well---it switches to a lightweight AB while still keeping high accuracy, comparable to MCDNN (seen on the demo site). Further, \name is always faster than MCDNN, while MCDNN, with a latency of 40-80 ms and even without switching overhead, has degraded performance under resource contention, and has to drop approximately every two out of three frames. As for the accuracy, there are only some occasional false classifications in \name (in total: 51 out of 3,000 frames, or 1.7\%). MCDNN, in this case, has slightly better accuracy (24 false classifications in 3,000 frames, or 0.8\%). We believe commonly used post-processing algorithms~\cite{kang2017t, han2016seq} can easily remove these occasional classification errors and both approaches can achieve very low inaccuracy.

\section{Discussion}
\label{sec_discussion}

\noindent \textbf{Training the approximation-enabled DNN} of \name may take longer than conventional DNNs, since at each iteration of training, different outports and input shapes try to minimize their own softmax loss, and thus, they may adjust internal weights of the DNN in conflicting ways. In our experiments with the VID dataset, we observe that our training time is around 3 days on our evaluation edge server, described in Section~\ref{sec_platform}, compared to 1 day to train a baseline ResNet-34 model. However, training being an offline process, the time is of less concern. However, training can be sped up by using one of various actively researched techniques for optimizing training, such as~\cite{le2011optimization}.

\noindent \textbf{Generalizing the approximation-enabled DNN to other architectures}. The shape and depth knobs are general to all CNN-based architectures. Theoretically, we can attach an outport (composed of SPP to adjust the shape and fully-connected layer to generate classification) to any layers. The legitimate input shape can be tricky and dependent on the specific architecture. Considering the training cost and exploration space, we select 7 shapes in multiples of 16 and 6 outports, in mostly equally spaced positions of the layers. More input shapes and outports enable finer-grained accuracy-latency trade-offs and the granularity of such a trade-off space depends on the design goal.

\section{Related Work}
\label{sec_related_work}

\noindent\textbf{System-wise optimization:}
There have been many optimization attempts to improve the efficiency of video analytics or other ML pipelines by building low power hardware and software accelerators for DNNs~\cite{deepx, reagen2016minerva, chen2017eyeriss, park2015big, han2016eie, parashar2017scnn, gao2017tetris, zhang2016cambricon} or improving application performance using database optimization, either on-premise~\cite{mahgoub2019sophia} or on cloud-hosted instances~\cite{mahgoub2020optimuscloud}. These are orthogonal and \name can also benefit from these optimizations. VideoStorm~\cite{videostorm}, Chameleon~\cite{jiang2018chameleon}, and Focus~\cite{hsieh2018focus} exploited various configurations and DNN models to handle video analytics queries in a situation-tailored manner. 
ExCamera~\cite{excamera-nsdi-2017} and Sonic~\cite{mahgoub2021sonic} enabled low-latency video processing on the cloud using serverless architecture (AWS Lambda~\cite{amazon_lambda}). Mainstream~\cite{jiang2018mainstream} proposed to share weights of DNNs across applications. These are all server-side solutions, requiring multiple models to be loaded simultaneously, which is challenging with resource constraints. NoScope~\cite{kang2017noscope} targeted to reduce the computation cost of video analytics queries on servers by leveraging a specialized model trained on a small subset of videos. Thus, its applicability is limited to a small subset of videos that the model has seen.
Closing-the-loop~\cite{xu2020closing} uses genetic algorithms to efficiently search for video editing parameters with lower computation cost. VideoChef~\cite{xu2018videochef} attempted to reduce the processing cost of video pipelines by dynamically changing approximation knobs of preprocessing filters in a content-aware manner. In contrast, \name, and concurrently developed ApproxDet~\cite{xu2020approxdet} (for video-specific object detection), approximate in the core DNN, which have a significantly different and computationally heavier program structure than filters. Due to this larger overhead of approximation in the core DNN, \name's adaptive tuning is challenging. Thus, we plan on using either distributed learning~\cite{ghoshal2015ensemble} or a reinforcement learning-based scheduler for refining this adaptive feature~\cite{thomas2018minerva}.

\noindent\textbf{DNN optimizations:}
Many solutions have been proposed to reduce computation cost of a DNN by controlling the precision of edge weights~\cite{hubara2017quantized, gupta2015deep, zhou2016dorefa, rastegari2016xnor} and restructuring or compressing a DNN model~\cite{denton2014exploiting, howard2017mobilenets, bhattacharya2016sparsification, iandola2016squeezenet, chen2015compressing, han2015learning, wen2016learning}. These are orthogonal to our work and \name's one-model approximation-enabled DNN can be further optimized by adopting such methods. There are several works that also present similar approximation knobs (input shape, outport depth). BranchyNet, CDL, and MSDNet~\cite{teerapittayanon2016branchynet, panda2016conditional, huang2017multi} propose early-exit branches in DNNs. However, BranchyNet and CDL only validate on small datasets like MNIST~\cite{MNIST} and CIFAR-10~\cite{CIFAR10} and have not shown practical techniques to selectively select these early-exit branches in an end-to-end pipeline. Such an adaptive system, in order to be useful (especially on embedded devices), needs to be responsive to resource contention, content characteristics, and users' requirement, \eg end-to-end latency SLA. MSDNet targets a very simple image classification task without a strong use case and does not show a data-driven manner of using the early exits. It would have been helpful to demonstrate the system's end-to-end latency on either a server-class or embedded device. BlockDrop~\cite{wu2018blockdrop} trains a policy network to determine whether to skip the execution of several residual blocks at inference time. However, its speedup is marginal and it cannot be applied directly to mobile devices for real-time classification.
\section{Conclusion}
\label{sec_conclusion}

There is a push to support streaming video analytics close to the source of the video, such as, IoT devices, surveillance cameras, or AR/VR gadgets. However, state-of-the-art heavy DNNs cannot run on such resource-constrained devices. Further, the runtime conditions for the DNN's execution may change due to changes in the resource availability on the device, the content characteristics, or user's requirements. Although several works create lightweight DNNs for resource-constrained clients, none of these can adapt to changing runtime conditions. We introduced \name, a video analytics system for embedded or mobile clients. It enables novel dynamic approximation techniques to achieve desired inference latency and accuracy trade-off under changing runtime conditions. It achieves this by enabling two approximation knobs within a single DNN model, rather than creating and maintaining an ensemble of models. It then estimates the effect on latency and accuracy due to changing content characteristics and changing levels of contention. We show that \name can adapt seamlessly at runtime to such changes to provide low and stable latency for object classification on a video stream. We quantitatively compare its performance to ResNet, MCDNN, MobileNets, NestDNN, and MSDNet, five state-of-the-art object classification DNNs.

\begin{acks}
This work is supported in part by the Wabash Heartland Innovation Network (WHIN) award from Lilly Endowment, NSF grant CCF-1919197, and Army Research Lab Contract number W911NF-20-2-0026. Any opinions, findings, and conclusions or recommendations expressed in this material are those of the authors and do not necessarily reflect the views of the sponsors.
\end{acks}

\bibliographystyle{ACM-Reference-Format}
\bibliography{main}


\begin{thebibliography}{87}


\ifx \showCODEN    \undefined \def \showCODEN     #1{\unskip}     \fi
\ifx \showDOI      \undefined \def \showDOI       #1{#1}\fi
\ifx \showISBNx    \undefined \def \showISBNx     #1{\unskip}     \fi
\ifx \showISBNxiii \undefined \def \showISBNxiii  #1{\unskip}     \fi
\ifx \showISSN     \undefined \def \showISSN      #1{\unskip}     \fi
\ifx \showLCCN     \undefined \def \showLCCN      #1{\unskip}     \fi
\ifx \shownote     \undefined \def \shownote      #1{#1}          \fi
\ifx \showarticletitle \undefined \def \showarticletitle #1{#1}   \fi
\ifx \showURL      \undefined \def \showURL       {\relax}        \fi
\providecommand\bibfield[2]{#2}
\providecommand\bibinfo[2]{#2}
\providecommand\natexlab[1]{#1}
\providecommand\showeprint[2][]{arXiv:#2}

\bibitem[\protect\citeauthoryear{Ausavarungnirun, Miller, Landgraf, Ghose,
  Gandhi, Jog, Rossbach, and Mutlu}{Ausavarungnirun et~al\mbox{.}}{2018}]%
        {ausavarungnirunASPLOS2018}
\bibfield{author}{\bibinfo{person}{Rachata Ausavarungnirun},
  \bibinfo{person}{Vance Miller}, \bibinfo{person}{Joshua Landgraf},
  \bibinfo{person}{Saugata Ghose}, \bibinfo{person}{Jayneel Gandhi},
  \bibinfo{person}{Adwait Jog}, \bibinfo{person}{Christopher~J Rossbach}, {and}
  \bibinfo{person}{Onur Mutlu}.} \bibinfo{year}{2018}\natexlab{}.
\newblock \showarticletitle{{Mask: Redesigning the GPU memory hierarchy to
  support multi-application concurrency}}. In \bibinfo{booktitle}{\emph{ACM
  SIGPLAN Notices}}, Vol.~\bibinfo{volume}{53}. ACM, \bibinfo{pages}{503--518}.
\newblock


\bibitem[\protect\citeauthoryear{Bagchi, Abdelzaher, Govindan, Shenoy, Atrey,
  Ghosh, and Xu}{Bagchi et~al\mbox{.}}{2020}]%
        {bagchi2020new}
\bibfield{author}{\bibinfo{person}{Saurabh Bagchi}, \bibinfo{person}{Tarek~F
  Abdelzaher}, \bibinfo{person}{Ramesh Govindan}, \bibinfo{person}{Prashant
  Shenoy}, \bibinfo{person}{Akanksha Atrey}, \bibinfo{person}{Pradipta Ghosh},
  {and} \bibinfo{person}{Ran Xu}.} \bibinfo{year}{2020}\natexlab{}.
\newblock \showarticletitle{New Frontiers in IoT: Networking, Systems,
  Reliability, and Security Challenges}.
\newblock \bibinfo{journal}{\emph{IEEE Internet of Things Journal}}
  \bibinfo{volume}{7}, \bibinfo{number}{12} (\bibinfo{year}{2020}),
  \bibinfo{pages}{11330--11346}.
\newblock


\bibitem[\protect\citeauthoryear{Bhattacharya and Lane}{Bhattacharya and
  Lane}{2016}]%
        {bhattacharya2016sparsification}
\bibfield{author}{\bibinfo{person}{Sourav Bhattacharya} {and}
  \bibinfo{person}{Nicholas~D Lane}.} \bibinfo{year}{2016}\natexlab{}.
\newblock \showarticletitle{Sparsification and separation of deep learning
  layers for constrained resource inference on wearables}. In
  \bibinfo{booktitle}{\emph{Proceedings of the 14th ACM Conference on Embedded
  Network Sensor Systems (Sensys)}}. ACM, \bibinfo{pages}{176--189}.
\newblock


\bibitem[\protect\citeauthoryear{Cardaci, Di~Ges{\`u}, Petrou, and
  Tabacchi}{Cardaci et~al\mbox{.}}{2009}]%
        {cardaci2009fuzzy}
\bibfield{author}{\bibinfo{person}{Maurizio Cardaci}, \bibinfo{person}{Vito
  Di~Ges{\`u}}, \bibinfo{person}{Maria Petrou}, {and}
  \bibinfo{person}{Marco~Elio Tabacchi}.} \bibinfo{year}{2009}\natexlab{}.
\newblock \showarticletitle{A fuzzy approach to the evaluation of image
  complexity}.
\newblock \bibinfo{journal}{\emph{Fuzzy Sets and Systems}}
  \bibinfo{volume}{160}, \bibinfo{number}{10} (\bibinfo{year}{2009}),
  \bibinfo{pages}{1474--1484}.
\newblock


\bibitem[\protect\citeauthoryear{Chen, Wilson, Tyree, Weinberger, and
  Chen}{Chen et~al\mbox{.}}{2015}]%
        {chen2015compressing}
\bibfield{author}{\bibinfo{person}{Wenlin Chen}, \bibinfo{person}{James
  Wilson}, \bibinfo{person}{Stephen Tyree}, \bibinfo{person}{Kilian
  Weinberger}, {and} \bibinfo{person}{Yixin Chen}.}
  \bibinfo{year}{2015}\natexlab{}.
\newblock \showarticletitle{Compressing neural networks with the hashing
  trick}. In \bibinfo{booktitle}{\emph{International Conference on Machine
  Learning}}. \bibinfo{pages}{2285--2294}.
\newblock


\bibitem[\protect\citeauthoryear{Chen, Krishna, Emer, and Sze}{Chen
  et~al\mbox{.}}{2017}]%
        {chen2017eyeriss}
\bibfield{author}{\bibinfo{person}{Yu-Hsin Chen}, \bibinfo{person}{Tushar
  Krishna}, \bibinfo{person}{Joel~S Emer}, {and} \bibinfo{person}{Vivienne
  Sze}.} \bibinfo{year}{2017}\natexlab{}.
\newblock \showarticletitle{Eyeriss: An energy-efficient reconfigurable
  accelerator for deep convolutional neural networks}.
\newblock \bibinfo{journal}{\emph{IEEE Journal of Solid-State Circuits}}
  \bibinfo{volume}{52}, \bibinfo{number}{1} (\bibinfo{year}{2017}),
  \bibinfo{pages}{127--138}.
\newblock


\bibitem[\protect\citeauthoryear{Corporation}{Corporation}{2018}]%
        {tx2}
\bibfield{author}{\bibinfo{person}{NVIDIA Corporation}.}
  \bibinfo{year}{2018}\natexlab{}.
\newblock \bibinfo{booktitle}{\emph{Jetson TX2 Module}}.
\newblock
\urldef\tempurl%
\url{https://developer.nvidia.com/embedded/buy/jetson-tx2}
\showURL{%
Retrieved May 5, 2020 from \tempurl}


\bibitem[\protect\citeauthoryear{Delimitrou and Kozyrakis}{Delimitrou and
  Kozyrakis}{2013}]%
        {delimitrou2013paragon}
\bibfield{author}{\bibinfo{person}{Christina Delimitrou} {and}
  \bibinfo{person}{Christos Kozyrakis}.} \bibinfo{year}{2013}\natexlab{}.
\newblock \showarticletitle{Paragon: QoS-aware scheduling for heterogeneous
  datacenters}.
\newblock \bibinfo{journal}{\emph{ACM SIGPLAN Notices}} \bibinfo{volume}{48},
  \bibinfo{number}{4} (\bibinfo{year}{2013}), \bibinfo{pages}{77--88}.
\newblock


\bibitem[\protect\citeauthoryear{Deng, Dong, Socher, Li, Li, and Fei-Fei}{Deng
  et~al\mbox{.}}{2009}]%
        {deng2009imagenet}
\bibfield{author}{\bibinfo{person}{Jia Deng}, \bibinfo{person}{Wei Dong},
  \bibinfo{person}{Richard Socher}, \bibinfo{person}{Li-Jia Li},
  \bibinfo{person}{Kai Li}, {and} \bibinfo{person}{Li Fei-Fei}.}
  \bibinfo{year}{2009}\natexlab{}.
\newblock \showarticletitle{Imagenet: A large-scale hierarchical image
  database}. In \bibinfo{booktitle}{\emph{Computer Vision and Pattern
  Recognition, 2009. CVPR 2009. IEEE Conference on}}. Ieee,
  \bibinfo{pages}{248--255}.
\newblock


\bibitem[\protect\citeauthoryear{Denton, Zaremba, Bruna, LeCun, and
  Fergus}{Denton et~al\mbox{.}}{2014}]%
        {denton2014exploiting}
\bibfield{author}{\bibinfo{person}{Emily~L Denton}, \bibinfo{person}{Wojciech
  Zaremba}, \bibinfo{person}{Joan Bruna}, \bibinfo{person}{Yann LeCun}, {and}
  \bibinfo{person}{Rob Fergus}.} \bibinfo{year}{2014}\natexlab{}.
\newblock \showarticletitle{Exploiting linear structure within convolutional
  networks for efficient evaluation}. In \bibinfo{booktitle}{\emph{Advances in
  neural information processing systems}}. \bibinfo{pages}{1269--1277}.
\newblock


\bibitem[\protect\citeauthoryear{Fang, Zeng, and Zhang}{Fang
  et~al\mbox{.}}{2018}]%
        {fang2018nestdnn}
\bibfield{author}{\bibinfo{person}{Biyi Fang}, \bibinfo{person}{Xiao Zeng},
  {and} \bibinfo{person}{Mi Zhang}.} \bibinfo{year}{2018}\natexlab{}.
\newblock \showarticletitle{Nestdnn: Resource-aware multi-tenant on-device deep
  learning for continuous mobile vision}. In
  \bibinfo{booktitle}{\emph{Proceedings of the 24th Annual International
  Conference on Mobile Computing and Networking}}. ACM,
  \bibinfo{pages}{115--127}.
\newblock


\bibitem[\protect\citeauthoryear{Fouladi, Wahby, Shacklett, Balasubramaniam,
  Zeng, Bhalerao, Sivaraman, Porter, and Winstein}{Fouladi
  et~al\mbox{.}}{2017}]%
        {excamera-nsdi-2017}
\bibfield{author}{\bibinfo{person}{Sadjad Fouladi}, \bibinfo{person}{Riad~S
  Wahby}, \bibinfo{person}{Brennan Shacklett}, \bibinfo{person}{Karthikeyan
  Balasubramaniam}, \bibinfo{person}{William Zeng}, \bibinfo{person}{Rahul
  Bhalerao}, \bibinfo{person}{Anirudh Sivaraman}, \bibinfo{person}{George
  Porter}, {and} \bibinfo{person}{Keith Winstein}.}
  \bibinfo{year}{2017}\natexlab{}.
\newblock \showarticletitle{Encoding, Fast and Slow: Low-Latency Video
  Processing Using Thousands of Tiny Threads.}. In
  \bibinfo{booktitle}{\emph{NSDI}}. \bibinfo{pages}{363--376}.
\newblock


\bibitem[\protect\citeauthoryear{Gao, Pu, Yang, Horowitz, and Kozyrakis}{Gao
  et~al\mbox{.}}{2017}]%
        {gao2017tetris}
\bibfield{author}{\bibinfo{person}{Mingyu Gao}, \bibinfo{person}{Jing Pu},
  \bibinfo{person}{Xuan Yang}, \bibinfo{person}{Mark Horowitz}, {and}
  \bibinfo{person}{Christos Kozyrakis}.} \bibinfo{year}{2017}\natexlab{}.
\newblock \showarticletitle{Tetris: Scalable and efficient neural network
  acceleration with 3d memory}.
\newblock \bibinfo{journal}{\emph{ACM SIGOPS Operating Systems Review}}
  \bibinfo{volume}{51}, \bibinfo{number}{2} (\bibinfo{year}{2017}),
  \bibinfo{pages}{751--764}.
\newblock


\bibitem[\protect\citeauthoryear{Ghoshal, Grama, Bagchi, and Chaterji}{Ghoshal
  et~al\mbox{.}}{2015}]%
        {ghoshal2015ensemble}
\bibfield{author}{\bibinfo{person}{Asish Ghoshal}, \bibinfo{person}{Ananth
  Grama}, \bibinfo{person}{Saurabh Bagchi}, {and} \bibinfo{person}{Somali
  Chaterji}.} \bibinfo{year}{2015}\natexlab{}.
\newblock \showarticletitle{An ensemble svm model for the accurate prediction
  of non-canonical microrna targets}. In \bibinfo{booktitle}{\emph{Proceedings
  of the 6th ACM Conference on Bioinformatics, Computational Biology and Health
  Informatics}}. \bibinfo{pages}{403--412}.
\newblock


\bibitem[\protect\citeauthoryear{Gupta, Agrawal, Gopalakrishnan, and
  Narayanan}{Gupta et~al\mbox{.}}{2015}]%
        {gupta2015deep}
\bibfield{author}{\bibinfo{person}{Suyog Gupta}, \bibinfo{person}{Ankur
  Agrawal}, \bibinfo{person}{Kailash Gopalakrishnan}, {and}
  \bibinfo{person}{Pritish Narayanan}.} \bibinfo{year}{2015}\natexlab{}.
\newblock \showarticletitle{Deep learning with limited numerical precision}. In
  \bibinfo{booktitle}{\emph{International Conference on Machine Learning}}.
  \bibinfo{pages}{1737--1746}.
\newblock


\bibitem[\protect\citeauthoryear{Han, Liu, Mao, Pu, Pedram, Horowitz, and
  Dally}{Han et~al\mbox{.}}{2016b}]%
        {han2016eie}
\bibfield{author}{\bibinfo{person}{Song Han}, \bibinfo{person}{Xingyu Liu},
  \bibinfo{person}{Huizi Mao}, \bibinfo{person}{Jing Pu},
  \bibinfo{person}{Ardavan Pedram}, \bibinfo{person}{Mark~A Horowitz}, {and}
  \bibinfo{person}{William~J Dally}.} \bibinfo{year}{2016}\natexlab{b}.
\newblock \showarticletitle{EIE: efficient inference engine on compressed deep
  neural network}. In \bibinfo{booktitle}{\emph{Computer Architecture (ISCA),
  2016 ACM/IEEE 43rd Annual International Symposium on}}. IEEE,
  \bibinfo{pages}{243--254}.
\newblock


\bibitem[\protect\citeauthoryear{Han, Pool, Tran, and Dally}{Han
  et~al\mbox{.}}{2015}]%
        {han2015learning}
\bibfield{author}{\bibinfo{person}{Song Han}, \bibinfo{person}{Jeff Pool},
  \bibinfo{person}{John Tran}, {and} \bibinfo{person}{William Dally}.}
  \bibinfo{year}{2015}\natexlab{}.
\newblock \showarticletitle{Learning both weights and connections for efficient
  neural network}. In \bibinfo{booktitle}{\emph{Advances in neural information
  processing systems}}. \bibinfo{pages}{1135--1143}.
\newblock


\bibitem[\protect\citeauthoryear{Han, Shen, Philipose, Agarwal, Wolman, and
  Krishnamurthy}{Han et~al\mbox{.}}{2016c}]%
        {mcdnn}
\bibfield{author}{\bibinfo{person}{Seungyeop Han}, \bibinfo{person}{Haichen
  Shen}, \bibinfo{person}{Matthai Philipose}, \bibinfo{person}{Sharad Agarwal},
  \bibinfo{person}{Alec Wolman}, {and} \bibinfo{person}{Arvind Krishnamurthy}.}
  \bibinfo{year}{2016}\natexlab{c}.
\newblock \showarticletitle{Mcdnn: An approximation-based execution framework
  for deep stream processing under resource constraints}. In
  \bibinfo{booktitle}{\emph{Proceedings of the 14th Annual International
  Conference on Mobile Systems, Applications, and Services}}. ACM,
  \bibinfo{pages}{123--136}.
\newblock


\bibitem[\protect\citeauthoryear{Han, Khorrami, Paine, Ramachandran,
  Babaeizadeh, Shi, Li, Yan, and Huang}{Han et~al\mbox{.}}{2016a}]%
        {han2016seq}
\bibfield{author}{\bibinfo{person}{Wei Han}, \bibinfo{person}{Pooya Khorrami},
  \bibinfo{person}{Tom~Le Paine}, \bibinfo{person}{Prajit Ramachandran},
  \bibinfo{person}{Mohammad Babaeizadeh}, \bibinfo{person}{Honghui Shi},
  \bibinfo{person}{Jianan Li}, \bibinfo{person}{Shuicheng Yan}, {and}
  \bibinfo{person}{Thomas~S Huang}.} \bibinfo{year}{2016}\natexlab{a}.
\newblock \showarticletitle{Seq-nms for video object detection}.
\newblock \bibinfo{journal}{\emph{arXiv preprint arXiv:1602.08465}}
  (\bibinfo{year}{2016}).
\newblock


\bibitem[\protect\citeauthoryear{Harris}{Harris}{2017}]%
        {unifor-mem}
\bibfield{author}{\bibinfo{person}{Mark Harris}.}
  \bibinfo{year}{2017}\natexlab{}.
\newblock \bibinfo{booktitle}{\emph{Unified Memory for CUDA Beginners}}.
\newblock
\urldef\tempurl%
\url{https://devblogs.nvidia.com/unified-memory-cuda-beginners/}
\showURL{%
Retrieved May 5, 2020 from \tempurl}


\bibitem[\protect\citeauthoryear{He, Zhang, Ren, and Sun}{He
  et~al\mbox{.}}{2014}]%
        {spp}
\bibfield{author}{\bibinfo{person}{Kaiming He}, \bibinfo{person}{Xiangyu
  Zhang}, \bibinfo{person}{Shaoqing Ren}, {and} \bibinfo{person}{Jian Sun}.}
  \bibinfo{year}{2014}\natexlab{}.
\newblock \showarticletitle{Spatial pyramid pooling in deep convolutional
  networks for visual recognition}. In \bibinfo{booktitle}{\emph{european
  conference on computer vision}}. Springer, \bibinfo{pages}{346--361}.
\newblock


\bibitem[\protect\citeauthoryear{He, Zhang, Ren, and Sun}{He
  et~al\mbox{.}}{2016}]%
        {resnet}
\bibfield{author}{\bibinfo{person}{Kaiming He}, \bibinfo{person}{Xiangyu
  Zhang}, \bibinfo{person}{Shaoqing Ren}, {and} \bibinfo{person}{Jian Sun}.}
  \bibinfo{year}{2016}\natexlab{}.
\newblock \showarticletitle{Deep residual learning for image recognition}. In
  \bibinfo{booktitle}{\emph{Proceedings of the IEEE conference on computer
  vision and pattern recognition}}. \bibinfo{pages}{770--778}.
\newblock


\bibitem[\protect\citeauthoryear{Howard, Zhu, Chen, Kalenichenko, Wang, Weyand,
  Andreetto, and Adam}{Howard et~al\mbox{.}}{2017}]%
        {howard2017mobilenets}
\bibfield{author}{\bibinfo{person}{Andrew~G Howard}, \bibinfo{person}{Menglong
  Zhu}, \bibinfo{person}{Bo Chen}, \bibinfo{person}{Dmitry Kalenichenko},
  \bibinfo{person}{Weijun Wang}, \bibinfo{person}{Tobias Weyand},
  \bibinfo{person}{Marco Andreetto}, {and} \bibinfo{person}{Hartwig Adam}.}
  \bibinfo{year}{2017}\natexlab{}.
\newblock \showarticletitle{Mobilenets: Efficient convolutional neural networks
  for mobile vision applications}.
\newblock \bibinfo{journal}{\emph{arXiv preprint arXiv:1704.04861}}
  (\bibinfo{year}{2017}).
\newblock


\bibitem[\protect\citeauthoryear{Hsieh, Ananthanarayanan, Bodik, Bahl,
  Philipose, Gibbons, and Mutlu}{Hsieh et~al\mbox{.}}{2018}]%
        {hsieh2018focus}
\bibfield{author}{\bibinfo{person}{Kevin Hsieh}, \bibinfo{person}{Ganesh
  Ananthanarayanan}, \bibinfo{person}{Peter Bodik}, \bibinfo{person}{Paramvir
  Bahl}, \bibinfo{person}{Matthai Philipose}, \bibinfo{person}{Phillip~B
  Gibbons}, {and} \bibinfo{person}{Onur Mutlu}.}
  \bibinfo{year}{2018}\natexlab{}.
\newblock \showarticletitle{Focus: Querying large video datasets with low
  latency and low cost}.
\newblock \bibinfo{journal}{\emph{arXiv preprint arXiv:1801.03493}}
  (\bibinfo{year}{2018}).
\newblock


\bibitem[\protect\citeauthoryear{Huang, Chen, Li, Wu, van~der Maaten, and
  Weinberger}{Huang et~al\mbox{.}}{2018}]%
        {huang2017multi}
\bibfield{author}{\bibinfo{person}{Gao Huang}, \bibinfo{person}{Danlu Chen},
  \bibinfo{person}{Tianhong Li}, \bibinfo{person}{Felix Wu},
  \bibinfo{person}{Laurens van~der Maaten}, {and} \bibinfo{person}{Kilian~Q
  Weinberger}.} \bibinfo{year}{2018}\natexlab{}.
\newblock \showarticletitle{Multi-scale dense networks for resource efficient
  image classification}. In \bibinfo{booktitle}{\emph{International Conference
  on Learning Representations (ICLR)}}.
\newblock


\bibitem[\protect\citeauthoryear{Huang, Liu, Van Der~Maaten, and
  Weinberger}{Huang et~al\mbox{.}}{2017}]%
        {huang2017densely}
\bibfield{author}{\bibinfo{person}{Gao Huang}, \bibinfo{person}{Zhuang Liu},
  \bibinfo{person}{Laurens Van Der~Maaten}, {and} \bibinfo{person}{Kilian~Q
  Weinberger}.} \bibinfo{year}{2017}\natexlab{}.
\newblock \showarticletitle{Densely connected convolutional networks}. In
  \bibinfo{booktitle}{\emph{Proceedings of the IEEE conference on computer
  vision and pattern recognition}}. \bibinfo{pages}{4700--4708}.
\newblock


\bibitem[\protect\citeauthoryear{Hubara, Courbariaux, Soudry, El-Yaniv, and
  Bengio}{Hubara et~al\mbox{.}}{2017}]%
        {hubara2017quantized}
\bibfield{author}{\bibinfo{person}{Itay Hubara}, \bibinfo{person}{Matthieu
  Courbariaux}, \bibinfo{person}{Daniel Soudry}, \bibinfo{person}{Ran
  El-Yaniv}, {and} \bibinfo{person}{Yoshua Bengio}.}
  \bibinfo{year}{2017}\natexlab{}.
\newblock \showarticletitle{Quantized Neural Networks: Training Neural Networks
  with Low Precision Weights and Activations.}
\newblock \bibinfo{journal}{\emph{Journal of Machine Learning Research}}
  \bibinfo{volume}{18} (\bibinfo{year}{2017}), \bibinfo{pages}{187--1}.
\newblock


\bibitem[\protect\citeauthoryear{Huynh, Lee, and Balan}{Huynh
  et~al\mbox{.}}{2017}]%
        {deepmon}
\bibfield{author}{\bibinfo{person}{Loc~N Huynh}, \bibinfo{person}{Youngki Lee},
  {and} \bibinfo{person}{Rajesh~Krishna Balan}.}
  \bibinfo{year}{2017}\natexlab{}.
\newblock \showarticletitle{Deepmon: Mobile gpu-based deep learning framework
  for continuous vision applications}. In \bibinfo{booktitle}{\emph{Proceedings
  of the 15th Annual International Conference on Mobile Systems, Applications,
  and Services}}. ACM, \bibinfo{pages}{82--95}.
\newblock


\bibitem[\protect\citeauthoryear{Iandola, Han, Moskewicz, Ashraf, Dally, and
  Keutzer}{Iandola et~al\mbox{.}}{2016}]%
        {iandola2016squeezenet}
\bibfield{author}{\bibinfo{person}{Forrest~N Iandola}, \bibinfo{person}{Song
  Han}, \bibinfo{person}{Matthew~W Moskewicz}, \bibinfo{person}{Khalid Ashraf},
  \bibinfo{person}{William~J Dally}, {and} \bibinfo{person}{Kurt Keutzer}.}
  \bibinfo{year}{2016}\natexlab{}.
\newblock \showarticletitle{Squeezenet: Alexnet-level accuracy with 50x fewer
  parameters and< 0.5 mb model size}.
\newblock \bibinfo{journal}{\emph{ICLR}} (\bibinfo{year}{2016}).
\newblock


\bibitem[\protect\citeauthoryear{Inc.}{Inc.}{2018}]%
        {amazon_lambda}
\bibfield{author}{\bibinfo{person}{Amazon Web~Services Inc.}}
  \bibinfo{year}{2018}\natexlab{}.
\newblock \bibinfo{booktitle}{\emph{AWS Lambda}}.
\newblock
\urldef\tempurl%
\url{https://aws.amazon.com/lambda/}
\showURL{%
Retrieved May 5, 2020 from \tempurl}


\bibitem[\protect\citeauthoryear{J{\"a}hne, Haussecker, and Geissler}{J{\"a}hne
  et~al\mbox{.}}{1999}]%
        {jahne1999handbook}
\bibfield{author}{\bibinfo{person}{Bernd J{\"a}hne}, \bibinfo{person}{Horst
  Haussecker}, {and} \bibinfo{person}{Peter Geissler}.}
  \bibinfo{year}{1999}\natexlab{}.
\newblock \bibinfo{booktitle}{\emph{Handbook of computer vision and
  applications}}. Vol.~\bibinfo{volume}{2}.
\newblock \bibinfo{publisher}{Citeseer}.
\newblock


\bibitem[\protect\citeauthoryear{Ji, Xu, Yang, and Yu}{Ji
  et~al\mbox{.}}{2013}]%
        {ji20133d}
\bibfield{author}{\bibinfo{person}{Shuiwang Ji}, \bibinfo{person}{Wei Xu},
  \bibinfo{person}{Ming Yang}, {and} \bibinfo{person}{Kai Yu}.}
  \bibinfo{year}{2013}\natexlab{}.
\newblock \showarticletitle{3D convolutional neural networks for human action
  recognition}.
\newblock \bibinfo{journal}{\emph{IEEE transactions on pattern analysis and
  machine intelligence}} \bibinfo{volume}{35}, \bibinfo{number}{1}
  (\bibinfo{year}{2013}), \bibinfo{pages}{221--231}.
\newblock


\bibitem[\protect\citeauthoryear{Jiang, Wong, Canel, Tang, Misra, Kaminsky,
  Kozuch, Pillai, Andersen, and Ganger}{Jiang et~al\mbox{.}}{2018b}]%
        {jiang2018mainstream}
\bibfield{author}{\bibinfo{person}{Angela~H Jiang}, \bibinfo{person}{Daniel L-K
  Wong}, \bibinfo{person}{Christopher Canel}, \bibinfo{person}{Lilia Tang},
  \bibinfo{person}{Ishan Misra}, \bibinfo{person}{Michael Kaminsky},
  \bibinfo{person}{Michael~A Kozuch}, \bibinfo{person}{Padmanabhan Pillai},
  \bibinfo{person}{David~G Andersen}, {and} \bibinfo{person}{Gregory~R
  Ganger}.} \bibinfo{year}{2018}\natexlab{b}.
\newblock \showarticletitle{Mainstream: Dynamic Stem-Sharing for Multi-Tenant
  Video Processing}. In \bibinfo{booktitle}{\emph{2018 USENIX Annual Technical
  Conference (USENIX ATC 18)}}. USENIX Association.
\newblock


\bibitem[\protect\citeauthoryear{Jiang, Ananthanarayanan, Bodik, Sen, and
  Stoica}{Jiang et~al\mbox{.}}{2018a}]%
        {jiang2018chameleon}
\bibfield{author}{\bibinfo{person}{Junchen Jiang}, \bibinfo{person}{Ganesh
  Ananthanarayanan}, \bibinfo{person}{Peter Bodik}, \bibinfo{person}{Siddhartha
  Sen}, {and} \bibinfo{person}{Ion Stoica}.} \bibinfo{year}{2018}\natexlab{a}.
\newblock \showarticletitle{Chameleon: scalable adaptation of video analytics}.
  In \bibinfo{booktitle}{\emph{Proceedings of the 2018 Conference of the ACM
  Special Interest Group on Data Communication}}. ACM,
  \bibinfo{pages}{253--266}.
\newblock


\bibitem[\protect\citeauthoryear{Kang, Emmons, Abuzaid, Bailis, and
  Zaharia}{Kang et~al\mbox{.}}{2017a}]%
        {kang2017noscope}
\bibfield{author}{\bibinfo{person}{Daniel Kang}, \bibinfo{person}{John Emmons},
  \bibinfo{person}{Firas Abuzaid}, \bibinfo{person}{Peter Bailis}, {and}
  \bibinfo{person}{Matei Zaharia}.} \bibinfo{year}{2017}\natexlab{a}.
\newblock \showarticletitle{NoScope: optimizing neural network queries over
  video at scale}.
\newblock \bibinfo{journal}{\emph{Proceedings of the VLDB Endowment}}
  \bibinfo{volume}{10}, \bibinfo{number}{11} (\bibinfo{year}{2017}),
  \bibinfo{pages}{1586--1597}.
\newblock


\bibitem[\protect\citeauthoryear{Kang, Li, Yan, Zeng, Yang, Xiao, Zhang, Wang,
  Wang, Wang, et~al\mbox{.}}{Kang et~al\mbox{.}}{2017b}]%
        {kang2017t}
\bibfield{author}{\bibinfo{person}{Kai Kang}, \bibinfo{person}{Hongsheng Li},
  \bibinfo{person}{Junjie Yan}, \bibinfo{person}{Xingyu Zeng},
  \bibinfo{person}{Bin Yang}, \bibinfo{person}{Tong Xiao},
  \bibinfo{person}{Cong Zhang}, \bibinfo{person}{Zhe Wang},
  \bibinfo{person}{Ruohui Wang}, \bibinfo{person}{Xiaogang Wang},
  {et~al\mbox{.}}} \bibinfo{year}{2017}\natexlab{b}.
\newblock \showarticletitle{T-CNN: Tubelets with convolutional neural networks
  for object detection from videos}.
\newblock \bibinfo{journal}{\emph{IEEE Transactions on Circuits and Systems for
  Video Technology}} (\bibinfo{year}{2017}).
\newblock


\bibitem[\protect\citeauthoryear{Kayiran, Nachiappan, Jog, Ausavarungnirun,
  Kandemir, Loh, Mutlu, and Das}{Kayiran et~al\mbox{.}}{2014}]%
        {kayiranMICRO2014}
\bibfield{author}{\bibinfo{person}{Onur Kayiran},
  \bibinfo{person}{Nachiappan~Chidambaram Nachiappan}, \bibinfo{person}{Adwait
  Jog}, \bibinfo{person}{Rachata Ausavarungnirun}, \bibinfo{person}{Mahmut~T
  Kandemir}, \bibinfo{person}{Gabriel~H Loh}, \bibinfo{person}{Onur Mutlu},
  {and} \bibinfo{person}{Chita~R Das}.} \bibinfo{year}{2014}\natexlab{}.
\newblock \showarticletitle{Managing GPU concurrency in heterogeneous
  architectures}. In \bibinfo{booktitle}{\emph{2014 47th Annual IEEE/ACM
  International Symposium on Microarchitecture}}. IEEE,
  \bibinfo{pages}{114--126}.
\newblock


\bibitem[\protect\citeauthoryear{Krizhevsky, Hinton, et~al\mbox{.}}{Krizhevsky
  et~al\mbox{.}}{2009}]%
        {CIFAR10}
\bibfield{author}{\bibinfo{person}{Alex Krizhevsky}, \bibinfo{person}{Geoffrey
  Hinton}, {et~al\mbox{.}}} \bibinfo{year}{2009}\natexlab{}.
\newblock \showarticletitle{Learning multiple layers of features from tiny
  images}.
\newblock  (\bibinfo{year}{2009}).
\newblock


\bibitem[\protect\citeauthoryear{Lane, Bhattacharya, Georgiev, Forlivesi, Jiao,
  Qendro, and Kawsar}{Lane et~al\mbox{.}}{2016}]%
        {deepx}
\bibfield{author}{\bibinfo{person}{Nicholas~D Lane}, \bibinfo{person}{Sourav
  Bhattacharya}, \bibinfo{person}{Petko Georgiev}, \bibinfo{person}{Claudio
  Forlivesi}, \bibinfo{person}{Lei Jiao}, \bibinfo{person}{Lorena Qendro},
  {and} \bibinfo{person}{Fahim Kawsar}.} \bibinfo{year}{2016}\natexlab{}.
\newblock \showarticletitle{Deepx: A software accelerator for low-power deep
  learning inference on mobile devices}. In
  \bibinfo{booktitle}{\emph{Proceedings of the 15th International Conference on
  Information Processing in Sensor Networks}}. IEEE Press, \bibinfo{pages}{23}.
\newblock


\bibitem[\protect\citeauthoryear{Laurenzano, Hill, Samadi, Mahlke, Mars, and
  Tang}{Laurenzano et~al\mbox{.}}{2016}]%
        {laurenzano2016input}
\bibfield{author}{\bibinfo{person}{Michael~A Laurenzano},
  \bibinfo{person}{Parker Hill}, \bibinfo{person}{Mehrzad Samadi},
  \bibinfo{person}{Scott Mahlke}, \bibinfo{person}{Jason Mars}, {and}
  \bibinfo{person}{Lingjia Tang}.} \bibinfo{year}{2016}\natexlab{}.
\newblock \showarticletitle{Input responsiveness: using canary inputs to
  dynamically steer approximation}.
\newblock \bibinfo{journal}{\emph{ACM SIGPLAN Notices}} \bibinfo{volume}{51},
  \bibinfo{number}{6} (\bibinfo{year}{2016}), \bibinfo{pages}{161--176}.
\newblock


\bibitem[\protect\citeauthoryear{Le, Ngiam, Coates, Lahiri, Prochnow, and
  Ng}{Le et~al\mbox{.}}{2011}]%
        {le2011optimization}
\bibfield{author}{\bibinfo{person}{Quoc~V Le}, \bibinfo{person}{Jiquan Ngiam},
  \bibinfo{person}{Adam Coates}, \bibinfo{person}{Abhik Lahiri},
  \bibinfo{person}{Bobby Prochnow}, {and} \bibinfo{person}{Andrew~Y Ng}.}
  \bibinfo{year}{2011}\natexlab{}.
\newblock \showarticletitle{On optimization methods for deep learning}. In
  \bibinfo{booktitle}{\emph{Proceedings of the 28th International Conference on
  International Conference on Machine Learning}}. Omnipress,
  \bibinfo{pages}{265--272}.
\newblock


\bibitem[\protect\citeauthoryear{LeCun, Bottou, Bengio, and Haffner}{LeCun
  et~al\mbox{.}}{1998}]%
        {MNIST}
\bibfield{author}{\bibinfo{person}{Yann LeCun}, \bibinfo{person}{L{\'e}on
  Bottou}, \bibinfo{person}{Yoshua Bengio}, {and} \bibinfo{person}{Patrick
  Haffner}.} \bibinfo{year}{1998}\natexlab{}.
\newblock \showarticletitle{Gradient-based learning applied to document
  recognition}.
\newblock \bibinfo{journal}{\emph{Proc. IEEE}} \bibinfo{volume}{86},
  \bibinfo{number}{11} (\bibinfo{year}{1998}), \bibinfo{pages}{2278--2324}.
\newblock


\bibitem[\protect\citeauthoryear{Liu, Shahroudy, Xu, and Wang}{Liu
  et~al\mbox{.}}{2016b}]%
        {liu2016spatio}
\bibfield{author}{\bibinfo{person}{Jun Liu}, \bibinfo{person}{Amir Shahroudy},
  \bibinfo{person}{Dong Xu}, {and} \bibinfo{person}{Gang Wang}.}
  \bibinfo{year}{2016}\natexlab{b}.
\newblock \showarticletitle{Spatio-temporal lstm with trust gates for 3d human
  action recognition}. In \bibinfo{booktitle}{\emph{European Conference on
  Computer Vision}}. Springer, \bibinfo{pages}{816--833}.
\newblock


\bibitem[\protect\citeauthoryear{Liu, Li, and Gruteser}{Liu
  et~al\mbox{.}}{2019}]%
        {liu2019edge}
\bibfield{author}{\bibinfo{person}{Luyang Liu}, \bibinfo{person}{Hongyu Li},
  {and} \bibinfo{person}{Marco Gruteser}.} \bibinfo{year}{2019}\natexlab{}.
\newblock \showarticletitle{Edge assisted real-time object detection for mobile
  augmented reality}.
\newblock  (\bibinfo{year}{2019}).
\newblock


\bibitem[\protect\citeauthoryear{Liu, Anguelov, Erhan, Szegedy, Reed, Fu, and
  Berg}{Liu et~al\mbox{.}}{2016a}]%
        {liu2016ssd}
\bibfield{author}{\bibinfo{person}{Wei Liu}, \bibinfo{person}{Dragomir
  Anguelov}, \bibinfo{person}{Dumitru Erhan}, \bibinfo{person}{Christian
  Szegedy}, \bibinfo{person}{Scott Reed}, \bibinfo{person}{Cheng-Yang Fu},
  {and} \bibinfo{person}{Alexander~C Berg}.} \bibinfo{year}{2016}\natexlab{a}.
\newblock \showarticletitle{Ssd: Single shot multibox detector}. In
  \bibinfo{booktitle}{\emph{European conference on computer vision}}. Springer,
  \bibinfo{pages}{21--37}.
\newblock


\bibitem[\protect\citeauthoryear{Lo, Cheng, Govindaraju, Ranganathan, and
  Kozyrakis}{Lo et~al\mbox{.}}{2015}]%
        {lo2015heracles}
\bibfield{author}{\bibinfo{person}{David Lo}, \bibinfo{person}{Liqun Cheng},
  \bibinfo{person}{Rama Govindaraju}, \bibinfo{person}{Parthasarathy
  Ranganathan}, {and} \bibinfo{person}{Christos Kozyrakis}.}
  \bibinfo{year}{2015}\natexlab{}.
\newblock \showarticletitle{Heracles: Improving resource efficiency at scale}.
  In \bibinfo{booktitle}{\emph{International Symposium on Computer Architecture
  (ISCA)}}, Vol.~\bibinfo{volume}{43}. ACM, \bibinfo{pages}{450--462}.
\newblock


\bibitem[\protect\citeauthoryear{Lu, Rallapalli, Chan, and La~Porta}{Lu
  et~al\mbox{.}}{2017}]%
        {lu2017modeling}
\bibfield{author}{\bibinfo{person}{Zongqing Lu}, \bibinfo{person}{Swati
  Rallapalli}, \bibinfo{person}{Kevin Chan}, {and} \bibinfo{person}{Thomas
  La~Porta}.} \bibinfo{year}{2017}\natexlab{}.
\newblock \showarticletitle{Modeling the resource requirements of convolutional
  neural networks on mobile devices}. In \bibinfo{booktitle}{\emph{Proceedings
  of the 25th ACM international conference on Multimedia}}. ACM,
  \bibinfo{pages}{1663--1671}.
\newblock


\bibitem[\protect\citeauthoryear{Mahgoub, Medoff, Kumar, Mitra, Klimovic,
  Chaterji, and Bagchi}{Mahgoub et~al\mbox{.}}{2020}]%
        {mahgoub2020optimuscloud}
\bibfield{author}{\bibinfo{person}{Ashraf Mahgoub},
  \bibinfo{person}{Alexander~Michaelson Medoff}, \bibinfo{person}{Rakesh
  Kumar}, \bibinfo{person}{Subrata Mitra}, \bibinfo{person}{Ana Klimovic},
  \bibinfo{person}{Somali Chaterji}, {and} \bibinfo{person}{Saurabh Bagchi}.}
  \bibinfo{year}{2020}\natexlab{}.
\newblock \showarticletitle{{OptimusCloud}: Heterogeneous Configuration
  Optimization for Distributed Databases in the Cloud}. In
  \bibinfo{booktitle}{\emph{2020 {USENIX} Annual Technical Conference ({USENIX
  ATC 20})}}. \bibinfo{pages}{189--203}.
\newblock


\bibitem[\protect\citeauthoryear{Mahgoub, Shankar, Mitra, Klimovic, Chaterji,
  and Bagchi}{Mahgoub et~al\mbox{.}}{2019}]%
        {mahgoub2019sophia}
\bibfield{author}{\bibinfo{person}{Ashraf Mahgoub}, \bibinfo{person}{Karthick
  Shankar}, \bibinfo{person}{Subrata Mitra}, \bibinfo{person}{Ana Klimovic},
  \bibinfo{person}{Somali Chaterji}, {and} \bibinfo{person}{Saurabh Bagchi}.}
  \bibinfo{year}{2019}\natexlab{}.
\newblock \showarticletitle{{Sophia}: Online reconfiguration of clustered nosql
  databases for time-varying workloads}. In \bibinfo{booktitle}{\emph{2021
  {USENIX} Annual Technical Conference ({USENIX ATC 19})}}.
  \bibinfo{pages}{223--240}.
\newblock


\bibitem[\protect\citeauthoryear{Mahgoub, Shankar, Mitra, Klimovic, Chaterji,
  and Bagchi}{Mahgoub et~al\mbox{.}}{2021}]%
        {mahgoub2021sonic}
\bibfield{author}{\bibinfo{person}{Ashraf Mahgoub}, \bibinfo{person}{Karthick
  Shankar}, \bibinfo{person}{Subrata Mitra}, \bibinfo{person}{Ana Klimovic},
  \bibinfo{person}{Somali Chaterji}, {and} \bibinfo{person}{Saurabh Bagchi}.}
  \bibinfo{year}{2021}\natexlab{}.
\newblock \showarticletitle{{SONIC}: Application-aware Data Passing for Chained
  Serverless Applications}. In \bibinfo{booktitle}{\emph{2021 {USENIX} Annual
  Technical Conference ({USENIX ATC 21})}}. \bibinfo{pages}{1--15}.
\newblock


\bibitem[\protect\citeauthoryear{Mario, Chacon, Alma, and Corral}{Mario
  et~al\mbox{.}}{2005}]%
        {edge-complexity-2}
\bibfield{author}{\bibinfo{person}{I Mario}, \bibinfo{person}{M Chacon},
  \bibinfo{person}{D Alma}, {and} \bibinfo{person}{S Corral}.}
  \bibinfo{year}{2005}\natexlab{}.
\newblock \showarticletitle{Image complexity measure: a human criterion free
  approach}. In \bibinfo{booktitle}{\emph{NAFIPS 2005-2005 Annual Meeting of
  the North American Fuzzy Information Processing Society}}. IEEE,
  \bibinfo{pages}{241--246}.
\newblock


\bibitem[\protect\citeauthoryear{Mars, Tang, Hundt, Skadron, and Soffa}{Mars
  et~al\mbox{.}}{2011}]%
        {mars2011bubble}
\bibfield{author}{\bibinfo{person}{Jason Mars}, \bibinfo{person}{Lingjia Tang},
  \bibinfo{person}{Robert Hundt}, \bibinfo{person}{Kevin Skadron}, {and}
  \bibinfo{person}{Mary~Lou Soffa}.} \bibinfo{year}{2011}\natexlab{}.
\newblock \showarticletitle{Bubble-up: Increasing utilization in modern
  warehouse scale computers via sensible co-locations}. In
  \bibinfo{booktitle}{\emph{Proceedings of the 44th annual IEEE/ACM
  International Symposium on Microarchitecture}}. ACM,
  \bibinfo{pages}{248--259}.
\newblock


\bibitem[\protect\citeauthoryear{Panda, Sengupta, and Roy}{Panda
  et~al\mbox{.}}{2016}]%
        {panda2016conditional}
\bibfield{author}{\bibinfo{person}{Priyadarshini Panda},
  \bibinfo{person}{Abhronil Sengupta}, {and} \bibinfo{person}{Kaushik Roy}.}
  \bibinfo{year}{2016}\natexlab{}.
\newblock \showarticletitle{Conditional deep learning for energy-efficient and
  enhanced pattern recognition}. In \bibinfo{booktitle}{\emph{2016 Design,
  Automation \& Test in Europe Conference \& Exhibition (DATE)}}. IEEE,
  \bibinfo{pages}{475--480}.
\newblock


\bibitem[\protect\citeauthoryear{Panta, Bagchi, and Midkiff}{Panta
  et~al\mbox{.}}{2011}]%
        {panta2011efficient}
\bibfield{author}{\bibinfo{person}{Rajesh~Krishna Panta},
  \bibinfo{person}{Saurabh Bagchi}, {and} \bibinfo{person}{Samuel~P Midkiff}.}
  \bibinfo{year}{2011}\natexlab{}.
\newblock \showarticletitle{Efficient incremental code update for sensor
  networks}.
\newblock \bibinfo{journal}{\emph{ACM Transactions on Sensor Networks (TOSN)}}
  \bibinfo{volume}{7}, \bibinfo{number}{4} (\bibinfo{year}{2011}),
  \bibinfo{pages}{1--32}.
\newblock


\bibitem[\protect\citeauthoryear{Parashar, Rhu, Mukkara, Puglielli, Venkatesan,
  Khailany, Emer, Keckler, and Dally}{Parashar et~al\mbox{.}}{2017}]%
        {parashar2017scnn}
\bibfield{author}{\bibinfo{person}{Angshuman Parashar}, \bibinfo{person}{Minsoo
  Rhu}, \bibinfo{person}{Anurag Mukkara}, \bibinfo{person}{Antonio Puglielli},
  \bibinfo{person}{Rangharajan Venkatesan}, \bibinfo{person}{Brucek Khailany},
  \bibinfo{person}{Joel Emer}, \bibinfo{person}{Stephen~W Keckler}, {and}
  \bibinfo{person}{William~J Dally}.} \bibinfo{year}{2017}\natexlab{}.
\newblock \showarticletitle{Scnn: An accelerator for compressed-sparse
  convolutional neural networks}. In \bibinfo{booktitle}{\emph{ACM SIGARCH
  Computer Architecture News}}, Vol.~\bibinfo{volume}{45}. ACM,
  \bibinfo{pages}{27--40}.
\newblock


\bibitem[\protect\citeauthoryear{Park, Kim, Kim, Kim, Kim, Yoon, and Yoo}{Park
  et~al\mbox{.}}{2015}]%
        {park2015big}
\bibfield{author}{\bibinfo{person}{Eunhyeok Park}, \bibinfo{person}{Dongyoung
  Kim}, \bibinfo{person}{Soobeom Kim}, \bibinfo{person}{Yong-Deok Kim},
  \bibinfo{person}{Gunhee Kim}, \bibinfo{person}{Sungroh Yoon}, {and}
  \bibinfo{person}{Sungjoo Yoo}.} \bibinfo{year}{2015}\natexlab{}.
\newblock \showarticletitle{Big/little deep neural network for ultra low power
  inference}. In \bibinfo{booktitle}{\emph{Proceedings of the 10th
  International Conference on Hardware/Software Codesign and System
  Synthesis}}. IEEE Press, \bibinfo{pages}{124--132}.
\newblock


\bibitem[\protect\citeauthoryear{Parkhi, Vedaldi, Zisserman,
  et~al\mbox{.}}{Parkhi et~al\mbox{.}}{2015}]%
        {parkhi2015deep}
\bibfield{author}{\bibinfo{person}{Omkar~M Parkhi}, \bibinfo{person}{Andrea
  Vedaldi}, \bibinfo{person}{Andrew Zisserman}, {et~al\mbox{.}}}
  \bibinfo{year}{2015}\natexlab{}.
\newblock \showarticletitle{Deep face recognition.}. In
  \bibinfo{booktitle}{\emph{BMVC}}, Vol.~\bibinfo{volume}{1}.
  \bibinfo{pages}{6}.
\newblock


\bibitem[\protect\citeauthoryear{Poppe}{Poppe}{2010}]%
        {poppe2010survey}
\bibfield{author}{\bibinfo{person}{Ronald Poppe}.}
  \bibinfo{year}{2010}\natexlab{}.
\newblock \showarticletitle{A survey on vision-based human action recognition}.
\newblock \bibinfo{journal}{\emph{Image and vision computing}}
  \bibinfo{volume}{28}, \bibinfo{number}{6} (\bibinfo{year}{2010}),
  \bibinfo{pages}{976--990}.
\newblock


\bibitem[\protect\citeauthoryear{Power}{Power}{2016}]%
        {YoutubeVideoLink}
\bibfield{author}{\bibinfo{person}{Canal~Max Power}.}
  \bibinfo{year}{2016}\natexlab{}.
\newblock \bibinfo{booktitle}{\emph{Sport Cars Drag Race Video}}.
\newblock
\urldef\tempurl%
\url{https://www.youtube.com/watch?v=Qj21A8HLQ0M}
\showURL{%
Retrieved May 5, 2020 from \tempurl}


\bibitem[\protect\citeauthoryear{Rastegari, Ordonez, Redmon, and
  Farhadi}{Rastegari et~al\mbox{.}}{2016}]%
        {rastegari2016xnor}
\bibfield{author}{\bibinfo{person}{Mohammad Rastegari},
  \bibinfo{person}{Vicente Ordonez}, \bibinfo{person}{Joseph Redmon}, {and}
  \bibinfo{person}{Ali Farhadi}.} \bibinfo{year}{2016}\natexlab{}.
\newblock \showarticletitle{Xnor-net: Imagenet classification using binary
  convolutional neural networks}. In \bibinfo{booktitle}{\emph{European
  Conference on Computer Vision}}. Springer, \bibinfo{pages}{525--542}.
\newblock


\bibitem[\protect\citeauthoryear{Reagen, Whatmough, Adolf, Rama, Lee, Lee,
  Hern{\'a}ndez-Lobato, Wei, and Brooks}{Reagen et~al\mbox{.}}{2016}]%
        {reagen2016minerva}
\bibfield{author}{\bibinfo{person}{Brandon Reagen}, \bibinfo{person}{Paul
  Whatmough}, \bibinfo{person}{Robert Adolf}, \bibinfo{person}{Saketh Rama},
  \bibinfo{person}{Hyunkwang Lee}, \bibinfo{person}{Sae~Kyu Lee},
  \bibinfo{person}{Jos{\'e}~Miguel Hern{\'a}ndez-Lobato},
  \bibinfo{person}{Gu-Yeon Wei}, {and} \bibinfo{person}{David Brooks}.}
  \bibinfo{year}{2016}\natexlab{}.
\newblock \showarticletitle{Minerva: Enabling low-power, highly-accurate deep
  neural network accelerators}. In \bibinfo{booktitle}{\emph{ACM SIGARCH
  Computer Architecture News}}, Vol.~\bibinfo{volume}{44}. IEEE Press,
  \bibinfo{pages}{267--278}.
\newblock


\bibitem[\protect\citeauthoryear{Redmon, Divvala, Girshick, and Farhadi}{Redmon
  et~al\mbox{.}}{2016}]%
        {redmon2016you}
\bibfield{author}{\bibinfo{person}{Joseph Redmon}, \bibinfo{person}{Santosh
  Divvala}, \bibinfo{person}{Ross Girshick}, {and} \bibinfo{person}{Ali
  Farhadi}.} \bibinfo{year}{2016}\natexlab{}.
\newblock \showarticletitle{You only look once: Unified, real-time object
  detection}. In \bibinfo{booktitle}{\emph{Proceedings of the IEEE conference
  on computer vision and pattern recognition}}. \bibinfo{pages}{779--788}.
\newblock


\bibitem[\protect\citeauthoryear{Ren, He, Girshick, and Sun}{Ren
  et~al\mbox{.}}{2015}]%
        {ren2015faster}
\bibfield{author}{\bibinfo{person}{Shaoqing Ren}, \bibinfo{person}{Kaiming He},
  \bibinfo{person}{Ross Girshick}, {and} \bibinfo{person}{Jian Sun}.}
  \bibinfo{year}{2015}\natexlab{}.
\newblock \showarticletitle{Faster r-cnn: Towards real-time object detection
  with region proposal networks}. In \bibinfo{booktitle}{\emph{Advances in
  neural information processing systems}}. \bibinfo{pages}{91--99}.
\newblock


\bibitem[\protect\citeauthoryear{Russakovsky, Deng, Su, Krause, Satheesh, Ma,
  Huang, Karpathy, Khosla, Bernstein, Berg, and Fei-Fei}{Russakovsky
  et~al\mbox{.}}{2015}]%
        {ILSVRC2015_VID}
\bibfield{author}{\bibinfo{person}{Olga Russakovsky}, \bibinfo{person}{Jia
  Deng}, \bibinfo{person}{Hao Su}, \bibinfo{person}{Jonathan Krause},
  \bibinfo{person}{Sanjeev Satheesh}, \bibinfo{person}{Sean Ma},
  \bibinfo{person}{Zhiheng Huang}, \bibinfo{person}{Andrej Karpathy},
  \bibinfo{person}{Aditya Khosla}, \bibinfo{person}{Michael Bernstein},
  \bibinfo{person}{Alexander~C. Berg}, {and} \bibinfo{person}{Li Fei-Fei}.}
  \bibinfo{year}{2015}\natexlab{}.
\newblock \showarticletitle{{ImageNet Large Scale Visual Recognition
  Challenge}}.
\newblock \bibinfo{journal}{\emph{International Journal of Computer Vision
  (IJCV)}} \bibinfo{volume}{115}, \bibinfo{number}{3} (\bibinfo{year}{2015}),
  \bibinfo{pages}{211--252}.
\newblock
\urldef\tempurl%
\url{https://doi.org/10.1007/s11263-015-0816-y}
\showDOI{\tempurl}


\bibitem[\protect\citeauthoryear{Schroff, Kalenichenko, and Philbin}{Schroff
  et~al\mbox{.}}{2015}]%
        {schroff2015facenet}
\bibfield{author}{\bibinfo{person}{Florian Schroff}, \bibinfo{person}{Dmitry
  Kalenichenko}, {and} \bibinfo{person}{James Philbin}.}
  \bibinfo{year}{2015}\natexlab{}.
\newblock \showarticletitle{Facenet: A unified embedding for face recognition
  and clustering}. In \bibinfo{booktitle}{\emph{Proceedings of the IEEE
  conference on computer vision and pattern recognition}}.
  \bibinfo{pages}{815--823}.
\newblock


\bibitem[\protect\citeauthoryear{Shankar, Wang, Xu, Mahgoub, and
  Chaterji}{Shankar et~al\mbox{.}}{2020}]%
        {shankar2020janus}
\bibfield{author}{\bibinfo{person}{Karthick Shankar},
  \bibinfo{person}{Pengcheng Wang}, \bibinfo{person}{Ran Xu},
  \bibinfo{person}{Ashraf Mahgoub}, {and} \bibinfo{person}{Somali Chaterji}.}
  \bibinfo{year}{2020}\natexlab{}.
\newblock \showarticletitle{JANUS: Benchmarking Commercial and Open-Source
  Cloud and Edge Platforms for Object and Anomaly Detection Workloads}. In
  \bibinfo{booktitle}{\emph{Proceedings of the IEEE International Conference on
  Cloud Computing}}. \bibinfo{pages}{1--9}.
\newblock


\bibitem[\protect\citeauthoryear{Simonyan and Zisserman}{Simonyan and
  Zisserman}{2014a}]%
        {simonyan2014two}
\bibfield{author}{\bibinfo{person}{Karen Simonyan} {and}
  \bibinfo{person}{Andrew Zisserman}.} \bibinfo{year}{2014}\natexlab{a}.
\newblock \showarticletitle{Two-stream convolutional networks for action
  recognition in videos}. In \bibinfo{booktitle}{\emph{Advances in neural
  information processing systems}}. \bibinfo{pages}{568--576}.
\newblock


\bibitem[\protect\citeauthoryear{Simonyan and Zisserman}{Simonyan and
  Zisserman}{2014b}]%
        {simonyan2014very}
\bibfield{author}{\bibinfo{person}{Karen Simonyan} {and}
  \bibinfo{person}{Andrew Zisserman}.} \bibinfo{year}{2014}\natexlab{b}.
\newblock \showarticletitle{Very deep convolutional networks for large-scale
  image recognition}.
\newblock \bibinfo{journal}{\emph{arXiv preprint arXiv:1409.1556}}
  (\bibinfo{year}{2014}).
\newblock


\bibitem[\protect\citeauthoryear{Szegedy, Liu, Jia, Sermanet, Reed, Anguelov,
  Erhan, Vanhoucke, and Rabinovich}{Szegedy et~al\mbox{.}}{2015}]%
        {szegedy2015going}
\bibfield{author}{\bibinfo{person}{Christian Szegedy}, \bibinfo{person}{Wei
  Liu}, \bibinfo{person}{Yangqing Jia}, \bibinfo{person}{Pierre Sermanet},
  \bibinfo{person}{Scott Reed}, \bibinfo{person}{Dragomir Anguelov},
  \bibinfo{person}{Dumitru Erhan}, \bibinfo{person}{Vincent Vanhoucke}, {and}
  \bibinfo{person}{Andrew Rabinovich}.} \bibinfo{year}{2015}\natexlab{}.
\newblock \showarticletitle{Going deeper with convolutions}. In
  \bibinfo{booktitle}{\emph{Proceedings of the IEEE conference on computer
  vision and pattern recognition}}. \bibinfo{pages}{1--9}.
\newblock


\bibitem[\protect\citeauthoryear{Taigman, Yang, Ranzato, and Wolf}{Taigman
  et~al\mbox{.}}{2014}]%
        {taigman2014deepface}
\bibfield{author}{\bibinfo{person}{Yaniv Taigman}, \bibinfo{person}{Ming Yang},
  \bibinfo{person}{Marc'Aurelio Ranzato}, {and} \bibinfo{person}{Lior Wolf}.}
  \bibinfo{year}{2014}\natexlab{}.
\newblock \showarticletitle{Deepface: Closing the gap to human-level
  performance in face verification}. In \bibinfo{booktitle}{\emph{Proceedings
  of the IEEE conference on computer vision and pattern recognition}}.
  \bibinfo{pages}{1701--1708}.
\newblock


\bibitem[\protect\citeauthoryear{Teerapittayanon, McDanel, and
  Kung}{Teerapittayanon et~al\mbox{.}}{2016}]%
        {teerapittayanon2016branchynet}
\bibfield{author}{\bibinfo{person}{Surat Teerapittayanon},
  \bibinfo{person}{Bradley McDanel}, {and} \bibinfo{person}{HT Kung}.}
  \bibinfo{year}{2016}\natexlab{}.
\newblock \showarticletitle{Branchynet: Fast inference via early exiting from
  deep neural networks}. In \bibinfo{booktitle}{\emph{2016 23rd International
  Conference on Pattern Recognition (ICPR)}}. IEEE,
  \bibinfo{pages}{2464--2469}.
\newblock


\bibitem[\protect\citeauthoryear{Thomas, Koo, Chaterji, and Bagchi}{Thomas
  et~al\mbox{.}}{2018}]%
        {thomas2018minerva}
\bibfield{author}{\bibinfo{person}{Tara~Elizabeth Thomas},
  \bibinfo{person}{Jinkyu Koo}, \bibinfo{person}{Somali Chaterji}, {and}
  \bibinfo{person}{Saurabh Bagchi}.} \bibinfo{year}{2018}\natexlab{}.
\newblock \showarticletitle{Minerva: A reinforcement learning-based technique
  for optimal scheduling and bottleneck detection in distributed factory
  operations}. In \bibinfo{booktitle}{\emph{2018 10th International Conference
  on Communication Systems \& Networks (COMSNETS)}}. IEEE,
  \bibinfo{pages}{129--136}.
\newblock


\bibitem[\protect\citeauthoryear{Wang, Li, and Ling}{Wang
  et~al\mbox{.}}{2018}]%
        {wang2018pelee}
\bibfield{author}{\bibinfo{person}{Robert~J Wang}, \bibinfo{person}{Xiang Li},
  {and} \bibinfo{person}{Charles~X Ling}.} \bibinfo{year}{2018}\natexlab{}.
\newblock \showarticletitle{Pelee: A real-time object detection system on
  mobile devices}. In \bibinfo{booktitle}{\emph{Advances in Neural Information
  Processing Systems}}. \bibinfo{pages}{1963--1972}.
\newblock


\bibitem[\protect\citeauthoryear{Wen, Wu, Wang, Chen, and Li}{Wen
  et~al\mbox{.}}{2016a}]%
        {wen2016learning}
\bibfield{author}{\bibinfo{person}{Wei Wen}, \bibinfo{person}{Chunpeng Wu},
  \bibinfo{person}{Yandan Wang}, \bibinfo{person}{Yiran Chen}, {and}
  \bibinfo{person}{Hai Li}.} \bibinfo{year}{2016}\natexlab{a}.
\newblock \showarticletitle{Learning structured sparsity in deep neural
  networks}. In \bibinfo{booktitle}{\emph{Advances in Neural Information
  Processing Systems}}. \bibinfo{pages}{2074--2082}.
\newblock


\bibitem[\protect\citeauthoryear{Wen, Zhang, Li, and Qiao}{Wen
  et~al\mbox{.}}{2016b}]%
        {wen2016discriminative}
\bibfield{author}{\bibinfo{person}{Yandong Wen}, \bibinfo{person}{Kaipeng
  Zhang}, \bibinfo{person}{Zhifeng Li}, {and} \bibinfo{person}{Yu Qiao}.}
  \bibinfo{year}{2016}\natexlab{b}.
\newblock \showarticletitle{A discriminative feature learning approach for deep
  face recognition}. In \bibinfo{booktitle}{\emph{European Conference on
  Computer Vision}}. Springer, \bibinfo{pages}{499--515}.
\newblock


\bibitem[\protect\citeauthoryear{Wu, Nagarajan, Kumar, Rennie, Davis, Grauman,
  and Feris}{Wu et~al\mbox{.}}{2018}]%
        {wu2018blockdrop}
\bibfield{author}{\bibinfo{person}{Zuxuan Wu}, \bibinfo{person}{Tushar
  Nagarajan}, \bibinfo{person}{Abhishek Kumar}, \bibinfo{person}{Steven
  Rennie}, \bibinfo{person}{Larry~S Davis}, \bibinfo{person}{Kristen Grauman},
  {and} \bibinfo{person}{Rogerio Feris}.} \bibinfo{year}{2018}\natexlab{}.
\newblock \showarticletitle{Blockdrop: Dynamic inference paths in residual
  networks}. In \bibinfo{booktitle}{\emph{Proceedings of the IEEE Conference on
  Computer Vision and Pattern Recognition}}. \bibinfo{pages}{8817--8826}.
\newblock


\bibitem[\protect\citeauthoryear{Xu, Koo, Kumar, Bai, Mitra, Misailovic, and
  Bagchi}{Xu et~al\mbox{.}}{2018a}]%
        {xu2018videochef}
\bibfield{author}{\bibinfo{person}{Ran Xu}, \bibinfo{person}{Jinkyu Koo},
  \bibinfo{person}{Rakesh Kumar}, \bibinfo{person}{Peter Bai},
  \bibinfo{person}{Subrata Mitra}, \bibinfo{person}{Sasa Misailovic}, {and}
  \bibinfo{person}{Saurabh Bagchi}.} \bibinfo{year}{2018}\natexlab{a}.
\newblock \showarticletitle{Videochef: efficient approximation for streaming
  video processing pipelines}. In \bibinfo{booktitle}{\emph{2018 USENIX Annual
  Technical Conference (USENIX ATC 18)}}. USENIX Association,
  \bibinfo{pages}{43--56}.
\newblock


\bibitem[\protect\citeauthoryear{Xu, Mitra, Rahman, Bai, Zhou, Bronevetsky, and
  Bagchi}{Xu et~al\mbox{.}}{2018b}]%
        {xu2018pythia}
\bibfield{author}{\bibinfo{person}{Ran Xu}, \bibinfo{person}{Subrata Mitra},
  \bibinfo{person}{Jason Rahman}, \bibinfo{person}{Peter Bai},
  \bibinfo{person}{Bowen Zhou}, \bibinfo{person}{Greg Bronevetsky}, {and}
  \bibinfo{person}{Saurabh Bagchi}.} \bibinfo{year}{2018}\natexlab{b}.
\newblock \showarticletitle{Pythia: Improving Datacenter Utilization via
  Precise Contention Prediction for Multiple Co-located Workloads}. In
  \bibinfo{booktitle}{\emph{Proceedings of the 19th International Middleware
  Conference}}. ACM, \bibinfo{pages}{146--160}.
\newblock


\bibitem[\protect\citeauthoryear{Xu, Wang, Petrangeli, Swaminathan, and
  Bagchi}{Xu et~al\mbox{.}}{2020a}]%
        {xu2020closing}
\bibfield{author}{\bibinfo{person}{Ran Xu}, \bibinfo{person}{Haoliang Wang},
  \bibinfo{person}{Stefano Petrangeli}, \bibinfo{person}{Viswanathan
  Swaminathan}, {and} \bibinfo{person}{Saurabh Bagchi}.}
  \bibinfo{year}{2020}\natexlab{a}.
\newblock \showarticletitle{Closing-the-Loop: A Data-Driven Framework for
  Effective Video Summarization}. In \bibinfo{booktitle}{\emph{Proceedings of
  the 22nd IEEE International Symposium on Multimedia (ISM)}}.
  \bibinfo{pages}{201--205}.
\newblock


\bibitem[\protect\citeauthoryear{Xu, Zhang, Wang, Lee, Mitra, Chaterji, Li, and
  Bagchi}{Xu et~al\mbox{.}}{2020b}]%
        {xu2020approxdet}
\bibfield{author}{\bibinfo{person}{Ran Xu}, \bibinfo{person}{Chen-lin Zhang},
  \bibinfo{person}{Pengcheng Wang}, \bibinfo{person}{Jayoung Lee},
  \bibinfo{person}{Subrata Mitra}, \bibinfo{person}{Somali Chaterji},
  \bibinfo{person}{Yin Li}, {and} \bibinfo{person}{Saurabh Bagchi}.}
  \bibinfo{year}{2020}\natexlab{b}.
\newblock \showarticletitle{ApproxDet: content and contention-aware approximate
  object detection for mobiles}. In \bibinfo{booktitle}{\emph{Proceedings of
  the 18th Conference on Embedded Networked Sensor Systems}}.
  \bibinfo{pages}{449--462}.
\newblock


\bibitem[\protect\citeauthoryear{Yang, Breslow, Mars, and Tang}{Yang
  et~al\mbox{.}}{2013}]%
        {yang2013bubble}
\bibfield{author}{\bibinfo{person}{Hailong Yang}, \bibinfo{person}{Alex
  Breslow}, \bibinfo{person}{Jason Mars}, {and} \bibinfo{person}{Lingjia
  Tang}.} \bibinfo{year}{2013}\natexlab{}.
\newblock \showarticletitle{Bubble-flux: Precise online qos management for
  increased utilization in warehouse scale computers}. In
  \bibinfo{booktitle}{\emph{International Symposium on Computer Architecture
  (ISCA)}}, Vol.~\bibinfo{volume}{41}. ACM, \bibinfo{pages}{607--618}.
\newblock


\bibitem[\protect\citeauthoryear{Yu and Winkler}{Yu and Winkler}{2013}]%
        {edge-compression-complexity}
\bibfield{author}{\bibinfo{person}{Honghai Yu} {and} \bibinfo{person}{Stefan
  Winkler}.} \bibinfo{year}{2013}\natexlab{}.
\newblock \showarticletitle{Image complexity and spatial information}. In
  \bibinfo{booktitle}{\emph{2013 Fifth International Workshop on Quality of
  Multimedia Experience (QoMEX)}}. IEEE, \bibinfo{pages}{12--17}.
\newblock


\bibitem[\protect\citeauthoryear{Zhang, Ananthanarayanan, Bodik, Philipose,
  Bahl, and Freedman}{Zhang et~al\mbox{.}}{2017}]%
        {videostorm}
\bibfield{author}{\bibinfo{person}{Haoyu Zhang}, \bibinfo{person}{Ganesh
  Ananthanarayanan}, \bibinfo{person}{Peter Bodik}, \bibinfo{person}{Matthai
  Philipose}, \bibinfo{person}{Paramvir Bahl}, {and} \bibinfo{person}{Michael~J
  Freedman}.} \bibinfo{year}{2017}\natexlab{}.
\newblock \showarticletitle{Live Video Analytics at Scale with Approximation
  and Delay-Tolerance.}. In \bibinfo{booktitle}{\emph{NSDI}},
  Vol.~\bibinfo{volume}{9}. \bibinfo{pages}{1}.
\newblock


\bibitem[\protect\citeauthoryear{Zhang, Du, Zhang, Lan, Liu, Li, Guo, Chen, and
  Chen}{Zhang et~al\mbox{.}}{2016}]%
        {zhang2016cambricon}
\bibfield{author}{\bibinfo{person}{Shijin Zhang}, \bibinfo{person}{Zidong Du},
  \bibinfo{person}{Lei Zhang}, \bibinfo{person}{Huiying Lan},
  \bibinfo{person}{Shaoli Liu}, \bibinfo{person}{Ling Li}, \bibinfo{person}{Qi
  Guo}, \bibinfo{person}{Tianshi Chen}, {and} \bibinfo{person}{Yunji Chen}.}
  \bibinfo{year}{2016}\natexlab{}.
\newblock \showarticletitle{Cambricon-x: An accelerator for sparse neural
  networks}. In \bibinfo{booktitle}{\emph{The 49th Annual IEEE/ACM
  International Symposium on Microarchitecture}}. IEEE Press,
  \bibinfo{pages}{20}.
\newblock


\bibitem[\protect\citeauthoryear{Zhang, Zhou, Lin, and Sun}{Zhang
  et~al\mbox{.}}{2018}]%
        {zhang2018shufflenet}
\bibfield{author}{\bibinfo{person}{Xiangyu Zhang}, \bibinfo{person}{Xinyu
  Zhou}, \bibinfo{person}{Mengxiao Lin}, {and} \bibinfo{person}{Jian Sun}.}
  \bibinfo{year}{2018}\natexlab{}.
\newblock \showarticletitle{Shufflenet: An extremely efficient convolutional
  neural network for mobile devices}. In \bibinfo{booktitle}{\emph{Proceedings
  of the IEEE Conference on Computer Vision and Pattern Recognition (CVPR)}}.
  \bibinfo{pages}{6848--6856}.
\newblock


\bibitem[\protect\citeauthoryear{Zhang, Laurenzano, Mars, and Tang}{Zhang
  et~al\mbox{.}}{2014}]%
        {zhang2014smite}
\bibfield{author}{\bibinfo{person}{Yunqi Zhang}, \bibinfo{person}{Michael~A
  Laurenzano}, \bibinfo{person}{Jason Mars}, {and} \bibinfo{person}{Lingjia
  Tang}.} \bibinfo{year}{2014}\natexlab{}.
\newblock \showarticletitle{Smite: Precise qos prediction on real-system smt
  processors to improve utilization in warehouse scale computers}. In
  \bibinfo{booktitle}{\emph{2014 47th Annual IEEE/ACM International Symposium
  on Microarchitecture}}. IEEE, \bibinfo{pages}{406--418}.
\newblock


\bibitem[\protect\citeauthoryear{Zhou, Wu, Ni, Zhou, Wen, and Zou}{Zhou
  et~al\mbox{.}}{2016}]%
        {zhou2016dorefa}
\bibfield{author}{\bibinfo{person}{Shuchang Zhou}, \bibinfo{person}{Yuxin Wu},
  \bibinfo{person}{Zekun Ni}, \bibinfo{person}{Xinyu Zhou}, \bibinfo{person}{He
  Wen}, {and} \bibinfo{person}{Yuheng Zou}.} \bibinfo{year}{2016}\natexlab{}.
\newblock \showarticletitle{Dorefa-net: Training low bitwidth convolutional
  neural networks with low bitwidth gradients}.
\newblock \bibinfo{journal}{\emph{arXiv preprint arXiv:1606.06160}}
  (\bibinfo{year}{2016}).
\newblock


\end{thebibliography}

\end{document}